\documentclass[10pt,twocolumn,letterpaper]{article}

\usepackage[pagenumbers]{iccv} %

\usepackage{multirow}
\usepackage[nolist,nohyperlinks]{acronym}
\usepackage{boldline}
\usepackage{bm}
\usepackage{pifont}

\newlength\savewidth

\usepackage{makecell}

\usepackage{bbding}
\usepackage{color}
\usepackage{pifont}
\usepackage{xcolor}
\usepackage{colortbl}

\usepackage{enumitem}
\setlist{nosep} %

\usepackage{caption}
\usepackage{subcaption}
\usepackage[skip=0.5ex]{subcaption}

\usepackage{xspace}

\newenvironment{fulljustify}
  {\par\setlength{\parfillskip}{0pt}} %
  {\par} %

\usepackage{arydshln}
\newcommand{\dashrule}[1][black]{%
  \color{#1}\rule[\dimexpr.5ex-.2pt]{4pt}{.4pt}\xleaders\hbox{\rule{4pt}{0pt}\rule[\dimexpr.5ex-.2pt]{4pt}{.4pt}}\hfill\kern0pt%
}
\newcommand{\rulecolor}[1]{%
  \def\CT@arc@{\color{#1}}%
}

\usepackage{capt-of}
\usepackage{duckuments}

\usepackage{algorithm}
\usepackage{algorithmic}

\usepackage{placeins}

\usepackage{amsmath}
\DeclareMathOperator*{\argmin}{\arg\!\min}
\usepackage{dsfont}

\newcommand{\davg}{\ensuremath{\delta_\text{avg}}\xspace}

\newcommand{\ua}{$\uparrow$}
\newcommand{\da}{$\downarrow$}
\newcommand{\B}[1]{\textbf{#1}}
\newcommand{\U}[1]{\underline{#1}}

\makeatletter
\renewcommand\paragraph{\@startsection{paragraph}{4}{\z@}{0.7ex plus 0.2ex minus 0.1ex}{-1em}{\normalfont\normalsize\bfseries}}
\makeatother

\newenvironment{tightalign}
  {\setlength{\abovedisplayskip}{3pt}%
   \setlength{\belowdisplayskip}{3pt}%
   \align}
  {\endalign}

\usepackage{textcomp}
\newcommand{\normaltilde}{\raisebox{0.5ex}{\texttildelow}}

\definecolor{iccvblue}{rgb}{0.21,0.49,0.74}
\usepackage[pagebackref,breaklinks,colorlinks,allcolors=iccvblue]{hyperref}

\def\paperID{426} %
\def\confName{ICCV}
\def\confYear{2025}

\title{Multi-View 3D Point Tracking}

\author{
    Frano Raji\v{c}$^{1}$ \quad
    Haofei Xu$^{1}$ \quad
    Marko Mihajlovic$^{1}$ \quad
    Siyuan Li$^{1}$ \quad
    Irem Demir$^{1}$ \\
    Emircan Gündoğdu$^{1}$ \hfill
    Lei Ke$^{2}$ \hfill
    Sergey Prokudin$^{1,3}$ \hfill
    Marc Pollefeys$^{1,4}$ \hfill
    Siyu Tang$^{1}$
\\[0.75ex]
    \small
    $^1$ETH Zürich \hfill
    $^2$Carnegie Mellon University \hfill
    $^3$Balgrist University Hospital \hfill
    $^4$Microsoft \\
}
 
\begin{document}

\twocolumn[{
\renewcommand\twocolumn[1][]{#1}
\maketitle
\begin{center}
    \vspace{-0.12cm}
    \captionsetup{type=figure}
    \includegraphics[width=\linewidth]{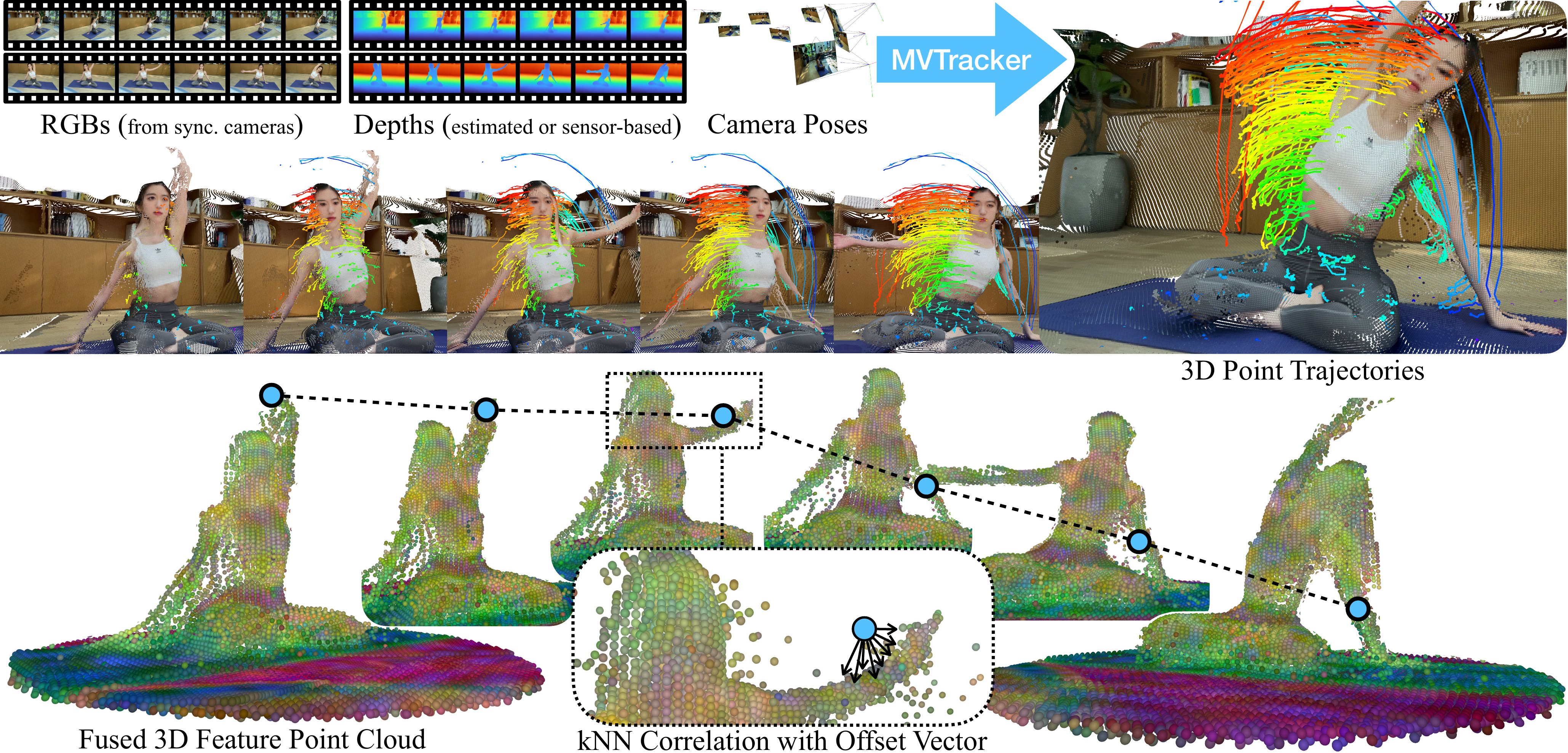}
    \vspace{-0.57cm}
    \captionof{figure}{
    We introduce \textbf{MVTracker}, the first \emph{data-driven multi-view 3D point tracker} for tracking arbitrary 3D points across multiple cameras. Our method fuses multi-view features into a unified 3D feature point cloud, within which it leverages kNN-based correlation to capture spatiotemporal relationships across views. A transformer then iteratively refines the point tracks, handling occlusions and adapting to varying camera setups without per-sequence optimization. This figure shows results on SelfCap~\cite{xu2024longvolcap_selfcap} using DUSt3R-based~\cite{wang2024dust3r} depth.
    } 
    \label{fig:teaser}
\end{center}
}]
\maketitle

\begin{abstract}
We introduce the first data-driven multi-view 3D point tracker, designed to track arbitrary points in dynamic scenes using multiple camera views. Unlike existing monocular trackers, which struggle with depth ambiguities and occlusion, or prior multi-camera methods that require over 20 cameras and tedious per-sequence optimization, our feed-forward model directly predicts 3D correspondences using a practical number of cameras (\eg, four), enabling robust and accurate online tracking. Given known camera poses and either sensor-based or estimated multi-view depth, our tracker fuses multi-view features into a unified point cloud and applies k-nearest-neighbors correlation alongside a transformer-based update to reliably estimate long-range 3D correspondences, even under occlusion. We train on 5K synthetic multi-view Kubric sequences and evaluate on two real-world benchmarks---Panoptic Studio and DexYCB---achieving median trajectory errors of 3.1\,cm and 2.0\,cm, respectively. Our method generalizes well to diverse camera setups of 1--8 views with varying vantage points and video lengths of 24--150 frames. By releasing our tracker alongside training and evaluation datasets, we aim to set a new standard for multi-view 3D tracking research and provide a practical tool for real-world applications. Project page: \url{https://ethz-vlg.github.io/mvtracker}.
\end{abstract}

\section{Introduction}
\label{sec:intro}
Tracking arbitrary points in 3D~\cite{vedula1999three} is a fundamental problem in computer vision, with numerous applications in dynamic scene reconstruction~\cite{jin2024stereo4d}, robotics~\cite{fang2023rh20t}, and augmented reality~\cite{pan2023aria}. While remarkable progress has been made with the recent advancement of 2D point tracking methods~\cite{karaev24cotracker3,karaev2023cotracker,harley2022particle,doersch2023tapvid,doersch2023tapir,doersch2024bootstap,cho2024local,zheng2023point}, they remain inherently limited in modeling 3D-consistent motion due to the fundamental ambiguity of the 3D-to-2D projection process. Thus, high-quality 3D point tracking remains a challenging task.

Scene flow approaches~\cite{vedula1999three,richardt2016dense,vogel20113d,teed2021raft} can estimate a dense 3D flow field for every 3D point, but they are usually limited to two consecutive video frames, whereas 3D point tracking operates on long sequences (\eg, tens or even hundreds of video frames). Recently, several methods~\cite{xiao2024spatialtracker,koppula2024tapvid,ngo2024delta} have been proposed to tackle the 3D point tracking task from monocular videos. However, their performance remains far from satisfactory for real-world applications that demand high quality and robustness, due to the difficulty of estimating 3D from single viewpoints in challenging scenarios such as occlusion and complex motion.

To address these challenges, we adopt a multi-camera setup and develop the first feed-forward model that can efficiently and robustly predict long-range 3D point trajectories from multi-view videos. Unlike previous multi-camera methods~\cite{vedula1999three,luiten2023dynamic,zheng2025gstar} that require more than 20 cameras and tedious per-sequence optimization, our approach enables feed-forward 3D point tracking with a practical and flexible number of cameras, such as a set of four camera views with arbitrary viewpoints. Thus, our method strikes a promising balance between accuracy and practicality, making it particularly suitable for real-world applications.

Key to our model is a dynamic fused 3D feature point cloud, constructed by combining the unprojected per-view depth maps, which not only effectively aggregates multi-view information to a global scene representation, but also facilitates reliable and efficient correspondence search with the k-nearest-neighbors (kNN) operation. This is different from the triplane representation used in the previous state-of-the-art method SpatialTracker~\cite{xiao2024spatialtracker}, which inevitably suffers from information loss during the triplane splatting process and is therefore less effective in processing varying numbers of input cameras. Moreover, our model performs reasonably well with different sources of depth input, whether from accurate depth sensors or noisy estimates from methods such as DUSt3R~\cite{wang2024dust3r} and VGGT~\cite{wang2025vggt}.

Using the fused dynamic point cloud, we retrieve local neighbors for each tracked point using a kNN search and compute multi-scale correlation features that capture both appearance similarity and 3D offset information. A spatiotemporal transformer then iteratively refines each point's 3D position and appearance over a sliding temporal window. Finally, the outputs from overlapping windows are merged to produce globally consistent 3D point trajectories. 

To train our model, we construct a synthetic multi-view dataset using Kubric~\cite{greff2022kubric} and simulate 5K sequences. We evaluate the model on two real-world datasets, DexYCB~\cite{dexycb} and Panoptic Studio~\cite{joo2015panoptic}, for which we construct 3D point trajectories by leveraging ground-truth object and hand pose estimation labels in DexYCB and merging existing monocular trajectory labels~\cite{koppula2024tapvid} in Panoptic Studio. We conduct extensive experiments on these multi-view video datasets, which suggest that our model outperforms baseline methods by a significant margin. Our model also performs well with different camera setups, different numbers of cameras, and different sources of depth maps, indicating the robustness of our proposed approach. We release our source code and models alongside the training and evaluation datasets to facilitate future research on multi-view 3D point tracking.

\section{Related Work}
\label{sec:related_work}

\paragraph{Scene Flow.}
Scene flow methods~\cite{vedula1999three,richardt2016dense,vogel20113d,teed2021raft} are usually designed to estimate the dense 3D motion between two consecutive video frames. Traditional methods~\cite{vedula1999three,richardt2016dense,vogel20113d} try to solve this task with an optimization framework, which is typically slow and requires many cameras (\eg, 31 in ~\cite{vedula1999three}) to well-regularize the optimization process. Modern approaches~\cite{yang2020upgrading,teed2021raft,liu2022camliflow} explore data-driven models to directly predict scene flow in a feed-forward manner. Despite recent progress, existing models are limited to two frames and are not able to track long-range correspondences in 3D.

\paragraph{2D Point Tracking.}
Recent years have witnessed significant progress in long-term 2D point tracking~\cite{doersch2023tapvid,harley2022particle,zheng2023point,wang2023omnimotion,doersch2023tapir,karaev2023cotracker,doersch2024bootstap,karaev24cotracker3,cho2024local,harley2025alltracker}. For instance, CoTracker2~\cite{karaev2023cotracker} leverages self-attention to aggregate spatial context from supporting tracks, while LocoTrack~\cite{cho2024local} extends traditional 2D correlation into bidirectional 4D correlation for more robust local matching. These methods typically estimate long-range point trajectories over an entire video, handling occlusions effectively. In this work, we investigate their extension to 3D, but with a sparse multi-view setup and known camera parameters, rather than monocular input. Moreover, unlike recent trackers~\cite{karaev24cotracker3,doersch2024bootstap} that reduce the synthetic-to-real domain gap via self-supervision or pseudolabeling, our approach focuses on direct supervision from synthetic data.

\paragraph{3D Point Tracking.}
3D point tracking has seen innovative contributions from recent methods~\cite{yu2023videodoodles,luiten2023dynamic,xiao2024spatialtracker,ngo2024delta,xiao2025spatialtrackerv2,tapip3d,som2024}. SpatialTracker~\cite{xiao2024spatialtracker} generalizes point tracking into 3D by integrating depth information through a Triplane representation~\cite{triplane} and DELTA~\cite{ngo2024delta} introduces a coarse-to-fine approach to estimate dense trajectories across the entire image plane, as opposed to sparse point tracking. Both methods assume monocular input. The concurrent SpatialTrackerV2~\cite{xiao2025spatialtrackerv2} and TAPIP3D~\cite{tapip3d} are also monocular. In a different approach, Dynamic 3DGS~\cite{luiten2023dynamic} leverages 3D Gaussian reconstructions over time to track dense scene elements in 3D and perform accurate 2D and 3D point tracking, but requires a multi-camera setup of 27 cameras in Panoptic Studio~\cite{joo2015panoptic} in addition to sparse depth from depth sensors for point cloud initialization and segmentation masks for relevant objects. We also use multiple input views but assume fewer cameras (\eg, four), which is more practical for real-world applications. Works such as~\cite{som2024,dreamscene4d,seidenschwarz2024dynomo} use monocular tracking priors and our experiments suggest that the multi-view extension of~\cite{som2024} is constrained by this. In contrast, our method directly learns a multi-view tracking prior that is neither Gaussian-based nor optimization-based. VideoDoodles~\cite{yu2023videodoodles} presents an interactive system that anchors hand-drawn animations to objects within a reconstructed 3D scene, harnessing optical flow and depth maps in a novel 3D point tracking algorithm. However, it is primarily designed for video editing rather than competitive performance on standard benchmarks such as TAPVid-2D~\cite{doersch2023tapvid}, whereas we focus on 3D point tracking accuracy.

\section{Method}
\label{sec:method}
\begin{figure*}
\begin{center}
    \captionsetup{type=figure}
    \includegraphics[width=\linewidth]{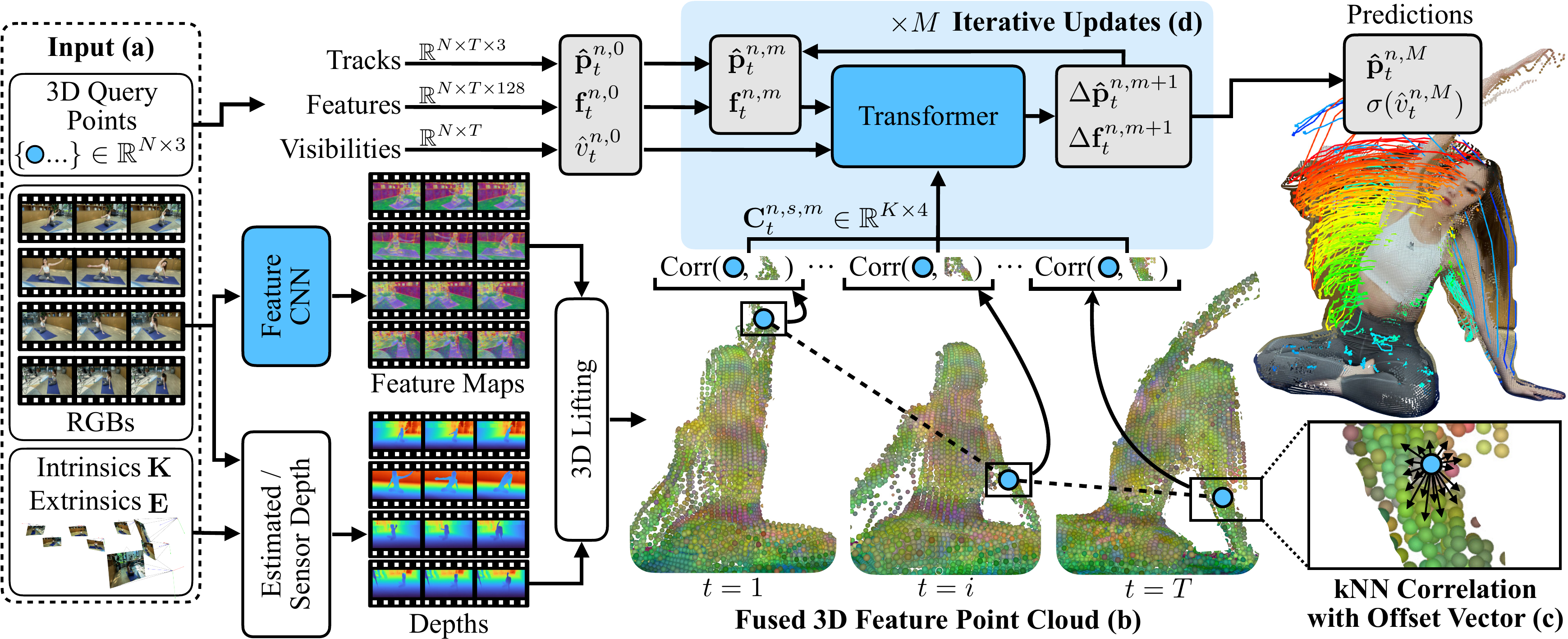}
    \vspace{-0.27in}
    \captionof{figure}{
    \textbf{MVTracker Pipeline}. 
    (a) Given synchronized multi-view RGB videos with known intrinsics and extrinsics, we extract per-view feature maps using a CNN-based encoder. (b) We construct a point cloud from estimated or sensor-based depth maps, associating each point with learned feature embeddings (\cref{sec:fused-point-cloud}). (c) We compute directed kNN-based correlation within the point cloud, capturing spatiotemporal relationships across views (\cref{sec:multi-scale-spatial-correlation}). (d) A transformer-based update module iteratively refines point trajectories using self-attention over multi-view feature correlations within a sliding temporal window (\cref{sec:Transformer}). (e) The model processes sequences in overlapping sliding windows, producing temporally consistent 3D point trajectories with occlusion-aware visibility predictions (\cref{sec:windowed-inference-and-unrolled-training}). The blue blocks denote trainable neural models. The visualized sequence is from SelfCap~\cite{xu2024longvolcap_selfcap} and uses DUSt3R~\cite{wang2024dust3r} to estimate depth.
    }
    \label{fig:pipeline}
\end{center}
\end{figure*}

Our goal is to perform online 3D point tracking across multiple views, given a set of synchronized RGB frames with known camera parameters. The input consists of video sequences captured from $V$ different camera views, with each frame containing a set of query points that need to be tracked over time. We estimate temporally consistent 3D trajectories for these points while also predicting visibility, handling occlusions, and adapting to scene dynamics. Our method runs online at 7.2 FPS given RGB-D input. For RGB-only input, we rely on external depth estimation (\eg, DUSt3R~\cite{wang2024dust3r} runs at 0.17 FPS and VGGT~\cite{wang2025vggt} at 3.1 FPS). An overview of our method is provided in \cref{fig:pipeline}.

Unlike single-view trackers that compute feature correlation on a 2D grid~\cite{karaev2023cotracker,karaev24cotracker3,cho2024local,ngo2024delta,doersch2023tapir,doersch2023tapvid,harley2022particle} or a triplane~\cite{xiao2024spatialtracker}, our method leverages a fused multi-view point cloud representation to establish $k$-nearest neighbors (kNN) feature correlations. While 2D-grid-based correlation can match irrelevant background pixels and triplane-based correlation inherently suffers from information loss, our kNN targets geometrically relevant 3D neighborhoods. This enables more robust tracking by integrating geometric consistency across views and learning motion priors directly from data.

\subsection{Problem Formulation}
\label{sec:problem-formulation}
Given a calibrated multi-view video sequence $\mathbf{I}_t^v \in \mathbb{R}^{H \times W \times 3}$ captured by $V$ cameras over $T$ frames, our goal is to track $N$ query points in 3D space. We denote each of the $N$ query points as 
$\mathbf{q}^n = (t_q^n,\, x_q^n,\, y_q^n,\, z_q^n) \in \mathbb{R}^4$, 
where $t_q^n$ denotes the query frame and $(x_q^n,\, y_q^n,\, z_q^n)$ its initial 3D location in world coordinates. Besides images and query points, the camera intrinsics $\mathbf{K}_t^v \in \mathbb{R}^{3 \times 3}$ and extrinsics $\mathbf{E}_t^v \in \mathbb{R}^{4 \times 4}$ are given for each view $v$ and frame~$t$.

The goal is to predict the 3D trajectory 
\{$\mathbf{p}_t^n=(x_t^n,\, y_t^n,\, z_t^n) \in \mathbb{R}^3$, $t \ge t_q^n$\}, along with visibilities 
\{$v_t^n \in \{\,0,1\,\}$\}
where \(v_t^n = 1\) denotes that the point is visible in at least one camera view, and \(v_t^n = 0\) indicates occlusion in all views. Note that the query location matches the ground-truth track location at time $t_q^n$, \ie $\mathbf{p}_{t_q^n}^n = (x_q^n,\, y_q^n,\, z_q^n)$.

\subsection{Point Cloud Encoding of Multi-view Videos}
\label{sec:fused-point-cloud}
\paragraph{Feature Maps.}
For each input frame $\mathbf{I}_t^v$, we extract $d$-dimensional per-view feature maps
$\Phi_t^v = \mathcal{\varphi}(\mathbf{I}_t^v)$
using a convolutional backbone (same as in \cite{karaev24cotracker3,karaev2023cotracker,xiao2024spatialtracker}).
We use a stride factor of $k=4$ to downscale the input resolution for computational efficiency, such that
$\Phi_t^v \in \mathbb{R}^{\frac{H}{k} \times \frac{W}{k} \times d}$.
We also compute the feature maps at $S=4$ different scales, \ie, $\Phi_t^{v,s} \in \mathbb{R}^{\frac{H}{k2^{s-1}} \times \frac{W}{k2^{s-1}} \times d}$, $s = 1,\dots,S$. These are obtained by downscaling the base features (\eg, via average pooling) and will later be used in our pyramid correlation.

\paragraph{Fused 3D Feature Point Cloud.}
We obtain multi-view-consistent depth maps $\mathbf{D}_t^v \in \mathbb{R}^{H \times W}$ for each view $v$ and frame $t$ using off-the-shelf depth estimation~\cite{wang2024dust3r,luiten2023dynamic,wang2025vggt} (see \cref{app:depth-analysis}). Sensor-based depth (\eg, Kinect) can also be used if available and preferred.  Using the depth and camera parameters, we lift each valid pixel $(u_x,u_y)$ into 3D:
\begin{tightalign}
    \label{eq:uplift}
    \mathbf{x} \;=\; \mathbf{E}_t^{v^{-1}}\bigl(\mathbf{K}_t^{v^{-1}}\left(\,u_x,u_y,1\,\right)^\top \cdot \mathbf{D}_t^v[\,u_y,u_x\,]\bigr).
\end{tightalign}
Invalid pixels are those that do not have a valid depth value due to missing Kinect depth values. For estimated depth, we retain all pixels and train for robustness (instead of thresholding by uncertainty predictions). Each of these lifted points $\mathbf{x}$ is associated with its corresponding feature from $\Phi_t^{v,s}$ and then fused across views into a single point cloud:
\begin{tightalign}
    \label{eq:pointcloud}
    \mathcal{X}_t^s \;=\; \bigl\{\bigl(\mathbf{x},\,\Phi_t^{v,s}[\,u_y,u_x\,]\bigr)\;\big|\;v&\in\{1,\dots,V\},\notag\\&\,(u_x,u_y)\in\Omega_t^v\bigr\},
\end{tightalign}
where $\Omega_t^v$ is the set of valid pixels in view $v$. This unified 3D representation facilitates the computation of spatial correlation across views, enabling multi-view-consistent tracking.

An alternative to our fused point cloud is to combine multi-view features via a single global \emph{triplane}~\cite{xiao2024spatialtracker}, projecting 3D points onto three orthogonal planes (\eg, XY, YZ, ZX). However, this inevitably suffers from projection collisions: different surfaces map to the same planar coordinates, causing destructive feature averaging. It also requires choosing a fixed bounding region for the scene, which leads to wasted resolution or clipped content for large or uncentered scenes. By contrast, our point cloud preserves features directly in 3D, avoids collisions, and adapts naturally to different scenes, resulting in more robust multi-view tracking.

\paragraph{Track Features.}
We assign a $d$-dimensional time-dependent appearance feature $\mathbf{f}_t^n$ to each tracked point. These features are initialized by sampling from the fused 3D feature point cloud at the query frame and 3D location:
\begin{tightalign}
\bigl(\mathbf{x}^{n,*},\mathbf{\phi}^{n,*}\bigr)
\;=\;\!\argmin_{\bigl(\mathbf{x},\,\mathbf{\phi}\bigr)\,\in\,\mathcal{X}_{\,t_n^q}^{\,1}}
\|\mathbf{x}-\mathbf{p}_{t_n^q}^n\|,
\end{tightalign}
for every track $n$.
We use the sampled feature \(\mathbf{\phi}^{n,*}\) to initialize \(\mathbf{f}_t^n\) for all \(t \ge t_n^q\). These track features are subsequently used when computing track-centric spatial correlation with nearest neighbors from the fused 3D feature point cloud (see \cref{sec:multi-scale-spatial-correlation}), but also refined by the spatiotemporal transformer to capture appearance changes over time (see \cref{sec:Transformer}).

\subsection{Multi-Scale Spatial Correlation}
\label{sec:multi-scale-spatial-correlation}
Instead of computing correlations separately for each view~\cite{karaev2023cotracker,karaev24cotracker3,cho2024local,ngo2024delta,doersch2023tapir,doersch2023tapvid,harley2022particle} or on an auxiliary triplane~\cite{xiao2024spatialtracker}, we establish correspondences directly in the fused feature point cloud using a multi-scale kNN approach. For each query point, we retrieve its $K$ nearest neighbors at different scales from $\mathcal{X}_t^s$ and compute local kNN feature correlations:
\begin{tightalign}
    \label{eq:correlation}
    \mathbf{C}_t^{n,s} \;=\; \bigl\{
    \langle \mathbf{f}_t^n,\mathbf{\phi}_k\rangle
    \;\big|\;
    \bigl(\mathbf{x}_k,\mathbf{\phi}_k\bigr)
    \in
    \mathcal{N}_K\!\bigl(\mathbf{\hat{p}}_t^{n},\mathcal{X}_t^s\bigr)
    \bigr\},
\end{tightalign}
where $\langle\cdot,\cdot\rangle$ is the dot product and $\mathcal{N}_K(\mathbf{\hat{p}}_t^{n},\mathcal{X}_t^s)$ finds the k-nearest neighbors around the current location estimate $\mathbf{\hat{p}}_t^{n}$ in the fused 3D feature point cloud. These correlations capture spatial dependencies at multiple scales and are fed into our transformer-based tracking module, allowing it to search increasingly larger 3D neighborhoods. For reference, average neighbor distances at the four scales on Panoptic Studio (made-up of \normaltilde10m–wide scenes) are $12.5 \pm 8.2$, $22.4 \pm 12.2$, $42.7 \pm 21.7$, and $85.8 \pm 44.0$\,cm. At the highest scale, this covers frame-to-frame motion of up to \normaltilde92~km/h at 30~FPS.

Unlike 2D correlation, where pixel offsets implicitly encode direction, our 3D point cloud representation requires explicit 3D offsets. Concretely, for each neighbor $(\mathbf{x}_k,\mathbf{\phi}_k)$ we concatenate the local similarity $\langle \mathbf{f}_t^n,\mathbf{\phi}_k\rangle$ with the offset vector $(\mathbf{x}_k-\mathbf{\hat{p}}_t^{n})$ to encode direction and distance from $\mathbf{\hat{p}}_t^{n}$. This offset helps the model disambiguate different nearest neighbors and enables correlation-based matching in 3D.

\subsection{Transformer-Based Iterative Tracking}
\label{sec:Transformer}
Our transformer refines track estimates over a sliding window of $T$ frames. For each track $n$ at time $t$, we construct a token $G_t^n=(\,\eta(\,\mathbf{\hat{p}}_t^{n}-\mathbf{\hat{p}}_{\,t_n^q}^{\,n}),\,\mathbf{f}_t^n,\,\mathbf{C}_t^{n,s},\,
\hat{v}_t^n)$, where $\eta(\cdot)$ is a sinusoidal positional encoding. The tokens $\{G_t^n\}_{t,n}$ are passed to a transformer $\mathcal{\psi}$, which applies temporal self-attention across time and cross-attention with a small set of learned virtual tracks to model spatial dependencies, following the design in CoTracker2~\cite{karaev24cotracker3}. The transformer outputs residual position and feature updates $(\Delta \mathbf{\hat{p}}_t^n,\Delta \mathbf{f}_t^n)=\mathcal{\psi}(G_t^n)$, which we apply iteratively for $m=1,\dots,M$:
\begin{tightalign}
\mathbf{\hat{p}}_t^{n,m+1} \;=\;\mathbf{\hat{p}}_t^{n,m} \,+\,\Delta\mathbf{\hat{p}}_t^{n,m+1}, \\
\mathbf{f}_t^{n,m+1} \;=\;\mathbf{f}_t^{n,m} \,+\,\Delta\mathbf{f}_t^{n,m+1}.
\end{tightalign}
After each iteration $m$, we recompute $\mathbf{C}_t^{n,s,m}$ based on the refined position and feature. At the final iteration $m=M$, we estimate visibility as $\hat{v}_t^{\,n}\;=\;\sigma\bigl(W\,\mathbf{f}_t^{n,M}\bigr)$,
where $\sigma(\cdot)$ is the sigmoid function and $W$ a learned projection matrix.

\subsection{Windowed Inference and Unrolled Training}
\label{sec:windowed-inference-and-unrolled-training}
To process long videos, we adopt a windowed approach as in CoTracker~\cite{karaev2023cotracker}. Let $T$ be the maximum window size; for a longer video of length $T'>T$, we divide it into $J$ overlapping windows of length $T$, each shifted by $T/2$ frames. Once the transformer completes $M$ iterative updates for window $j$, its final track estimates initialize window $j+1$, allowing refined trajectories to propagate. Let $\mathbf{\hat{p}}_t^{\,n,m,j}$  be the predicted location of query $n$ at time $t$ after iteration $m$ in window $j$, and $\hat{v}_t^{\,n,j}$ its visibility (predicted only at the final iteration). During training, we unroll these windowed updates so that each iteration learns to correct or refine predictions from previous windows and iterations.

\subsection{Supervision}
Training is supervised with ground-truth trajectories $\mathbf{p}_t^n$ and visibility labels $v_t^n$. The total loss consists of a position loss ($\mathcal{L}_\text{xyz}$) and a visibility loss ($\mathcal{L}_\text{vis}$), balanced by $\lambda_\text{vis}$:
\begin{tightalign}
\mathcal{L}
    =
    \mathcal{L}_\text{xyz}
    + \lambda_{\text{vis}} \mathcal{L}_{\text{vis}}.
\end{tightalign}
In particular, the position loss is the weighted $\ell_1$-norm error between the predicted and the ground-truth 3D positions:
\begin{tightalign}
\mathcal{L}_{\text{xyz}}
    \;=\;
    \sum_{j,m,n,t}
    \frac{\gamma^{\,M-m}}{JMNT}
    \;
    \bigl\|\, \mathbf{\hat{p}}_t^{\,n,m,j} - \mathbf{p}_t^{\,n,m,j}\bigr\|_1
     \,,
\end{tightalign}
where $j$ indexes the windows $J$, $m$ the iterative updates $M$, $n$ the trajectories $N$, $t$ the number of frames $T$, and $\gamma$ is a weighting factor used to penalize the later iterations of the iterative refinement more. The visibility loss addresses class imbalance by minimizing the balanced binary cross-entropy (B-BCE) between predicted and ground-truth labels:
\begin{tightalign}
\mathcal{L}_{\text{vis}}
    \;=\;
    \sum_{j,t,n}
    \frac{1}{J\,T\,N}
    \operatorname{B{-}BCE}\bigl(\hat{v}_t^{\,n,j},\,v_t^{\,n,j}\bigr).
\end{tightalign}

\section{Experiments}
\label{sec:experiments}

\paragraph{Datasets.}
To train our model, we generate a synthetic multi-view dataset, MV-Kub, using Kubric~\cite{greff2022kubric}, simulating 5K multi-view video sequences. Since no large-scale datasets exist for multi-view 3D point tracking, we rely on synthetic data for supervised training. For evaluation on the hand dexterity dataset DexYCB~\cite{dexycb}, we sample 10 scenes, leveraging ground-truth object and hand poses to generate track labels. We also construct a Panoptic Studio evaluation dataset by merging existing monocular labels from TAPVid-3D~\cite{koppula2024tapvid} for a total of 6 scenes. Evaluation query points are randomly sampled from both static and moving surfaces.

All results, unless otherwise stated, are reported assuming the availability of estimated depth~\cite{wang2024dust3r} for DexYCB, ground-truth optimization-based depth~\cite{luiten2023dynamic} for Panoptic Studio, and simulated depth for MV-Kub. In \cref{app:sub:ablation} we ablate the impact of depth sources on the performance. 

\begin{table*}
\setlength{\tabcolsep}{2pt}
\centering
\caption{\textbf{Quantitative evaluation of multi-view 3D point tracking performance.} We report results on Panoptic Studio~\cite{koppula2024tapvid}, DexYCB~\cite{dexycb}, and Multi-View Kubric~\cite{greff2022kubric}. Our method outperforms existing approaches across all datasets, achieving the highest accuracy (\davg) and occlusion-aware tracking performance (AJ), while maintaining lower median trajectory error (MTE). Notably, our model surpasses the triplane-based baseline, indicating the effectiveness of kNN-directed correlation for multi-view tracking. We report MTE in centimeters. F denotes that the method is feed-forward, data-driven, and online. M denotes that the method supports fusing multi-view inputs. We use estimated depth~\cite{wang2024dust3r} for DexYCB, ground-truth optimization-based depth~\cite{luiten2023dynamic} for Panoptic Studio, and simulated depth for Kubric.}
\vspace{-0.3cm}
\begin{tabular}{llccc ccccc ccccc cccc}
\toprule
  \multirow{2}{*}[-0.2em]{Method}
& \multirow{2}{*}[-0.2em]{Train}
& \multirow{2}{*}[-0.2em]{F}
& \multirow{2}{*}[-0.2em]{M}
& \xspace
& \multicolumn{4}{c}{Panoptic Studio~\cite{koppula2024tapvid}}
& \xspace
& \multicolumn{4}{c}{DexYCB~\cite{dexycb} (\scriptsize DUSt3R depth\normalsize)}
& \xspace
& \multicolumn{4}{c}{Multi-View Kubric~\cite{greff2022kubric}}
\\ \cmidrule(lr){6-9} \cmidrule(lr){11-14} \cmidrule(lr){16-19}
                                             &           &           &           &&   AJ \ua & \davg\ua &   OA \ua &  MTE \da &&   AJ \ua & \davg\ua &   OA \ua &  MTE \da &&   AJ \ua & \davg\ua &   OA \ua &  MTE \da \\ \hline
Dynamic 3DGS~\cite{luiten2023dynamic}        & \ding{55} & \ding{55} & \ding{51} &&    66.5  &    92.4  &    74.1  &     3.9  &&    45.7  &    57.1  &    81.3  &    11.3  &&    30.4  &    48.6  &    68.3  &    11.2  \\
Shape of Motion~\cite{som2024}               & \ding{55} & \ding{55} & \ding{51} &&    72.6  &    89.2  &    82.1  &     4.8  &&    36.2  &    53.6  &    63.3  &     8.0  &&    57.8  &    63.4  &    86.2  &     5.3  \\
SceneTracker~\cite{wang2024scenetracker}     &      LSFO & \ding{51} & \ding{55} &&      --  &    85.1  &      --  &     6.3  &&      --  &    61.1  &      --  &     5.7  &&      --  &    56.9  &      --  &     7.8  \\ 
LocoTrack~\cite{cho2024local}                &       Kub & \ding{51} & \ding{55} &&    65.8  &    79.4  &    79.6  &    11.8  &&    27.8  &    38.6  &    77.0  &    22.8  &&    52.5  &    58.4  &    76.5  &    12.9  \\ 
DELTA~\cite{ngo2024delta}                    &       Kub & \ding{51} & \ding{55} &&    68.1  &    86.3  &    79.2  &     5.9  &&    36.8  &    51.6  &    61.0  &    18.3  &&    57.4  &    68.4  &    77.9  &     9.6  \\ 
CoTracker2~\cite{karaev2023cotracker}        &       Kub & \ding{51} & \ding{55} &&    69.5  &    83.2  &    80.7  &    10.4  &&    28.8  &    40.5  &    76.2  &    20.8  &&    54.6  &    60.3  &    79.8  &    10.4  \\ 
CoTracker3~\cite{karaev24cotracker3}         &   Kub+15k & \ding{51} & \ding{55} &&    74.5  &    84.5  &    86.4  &     8.6  &&    29.4  &    40.3  &    78.6  &    22.0  &&    55.1  &    60.6  &    77.7  &    11.9  \\ 
SpaTracker~\cite{xiao2024spatialtracker}     &       Kub & \ding{51} & \ding{55} &&    55.5  &    70.8  &    81.7  &     9.9  &&    48.1  &    62.6  &    73.9  &     5.5  &&    43.6  &    55.1  &    76.3  &     5.1  \\ 
SpaTracker~\cite{xiao2024spatialtracker}     &    MV-Kub & \ding{51} & \ding{55} &&    61.5  &    82.3  &    75.5  &     7.3  &&    58.3  &    72.0  &    80.2  &     5.9  &&    65.5  &    77.6  &    83.1  &     2.2  \\ 
SpaTrackerV2~\cite{xiao2025spatialtrackerv2}
                   & See~\cite{xiao2025spatialtrackerv2} & \ding{51} & \ding{55} &&    75.3  &    85.1  &    91.2  &     6.9  &&    35.5  &    45.1  &    91.3  &     9.8  &&    58.6  &    69.3  &    86.3  &     3.9  \\ 
TAPIP3D~\cite{tapip3d}                       &       Kub & \ding{51} & \ding{55} &&    84.3  &    93.8  &    91.4  &  \B{3.1} &&    38.8  &    50.0  &    90.1  &     8.2  &&    72.4  &    86.5  &    85.4  &     1.3  \\ 
Triplane Baseline                            &    MV-Kub & \ding{51} & \ding{51} &&    65.1  &    81.8  &    82.8  &     7.2  &&    57.5  &    71.0  &    81.3  &     4.3  &&    74.7  &    85.2  &    90.4  &     1.2  \\ 

\rowcolor{cyan!16} MVTracker (ours)          &    MV-Kub & \ding{51} & \ding{51} && \B{86.0} & \B{94.7} & \B{92.3} & \B{3.1} && \B{71.6} & \B{80.6} & \B{91.3} &  \B{2.0} && \B{81.4} & \B{90.0} & \B{93.7} & \B{0.7}  \\ 
\bottomrule
\end{tabular}
\label{tab:main-results-table}
\end{table*}

\paragraph{Evaluation Metrics.}
As no standard metrics exist for multi-view 3D point tracking, we extend those from monocular benchmarks~\cite{doersch2023tapvid}, reporting four key metrics: (1) Median Trajectory Error (MTE) for visible points, quantifying spatial accuracy; (2) \davg, the average location accuracy over a set of threshold distances; (3) Occlusion Accuracy (OA), measuring the model's binary visibility prediction across views; and (4) Average Jaccard (AJ), which jointly evaluates occlusion and position accuracy. All metrics are computed per track, averaged within a scene, and finally aggregated across all dataset scenes. See \cref{app:subsec:metrics} for details. 

\paragraph{Training.} 
We train our method on MV-Kub for 200K steps on 8 GH200 chips with 96GB GPU memory over 8 days, with a batch size of 8 and up to 2048 trajectories per sample. Our model is implemented in PyTorch (with Lightning Fabric for multi-node training) and optimized using AdamW \cite{loshchilov2017decoupled}. The model operates with a latent feature dimension of $d=128$ and is trained on MV-Kub sequences of up to 24 frames using a sliding window of 12 frames. We use six-head self-attention layers with a hidden size of 256. We apply a range of augmentations to improve generalization and robustness. See \cref{app:sec:trainingdetails} for further details. 

\paragraph{Baselines.}
Since there currently exist no other feed-forward multi-view point trackers, we implement a triplane-based baseline by extending SpaTracker~\cite{xiao2024spatialtracker} to fuse multi-view features into a single global triplane, referred to as the ``Triplane Baseline'' in experiments. 
We also evaluate against two multi-view methods that require test-time optimization: Shape of Motion~\cite{som2024} and Dynamic 3DGS~\cite{luiten2023dynamic}; please refer to \cref{app:sec:baseline_implementation} for their implementation details. 

We further evaluate against monocular 2D and 3D point trackers. For 2D point trackers (CoTracker2~\cite{karaev2023cotracker}, CoTracker3~\cite{karaev24cotracker3}, LocoTrack~\cite{cho2024local}), we run the models on each camera separately with the query points that are deemed to be best visible from that view (based on their projected depth), then lift the resulting 2D estimates to 3D using the known intrinsics and the same depth used by our model. 
We similarly run 3D trackers (DELTA~\cite{ngo2024delta}, SpaTracker~\cite{xiao2024spatialtracker}, SpaTrackerV2~\cite{xiao2025spatialtrackerv2}, TAPIP3D~\cite{tapip3d}), but give depth as input to directly get 3D tracks. We finally merge per-camera predictions into the world space using known camera poses. 

\subsection{Method Comparisons} \label{sec:results}

\begin{figure*}
    \centering
    \includegraphics[width=\linewidth]{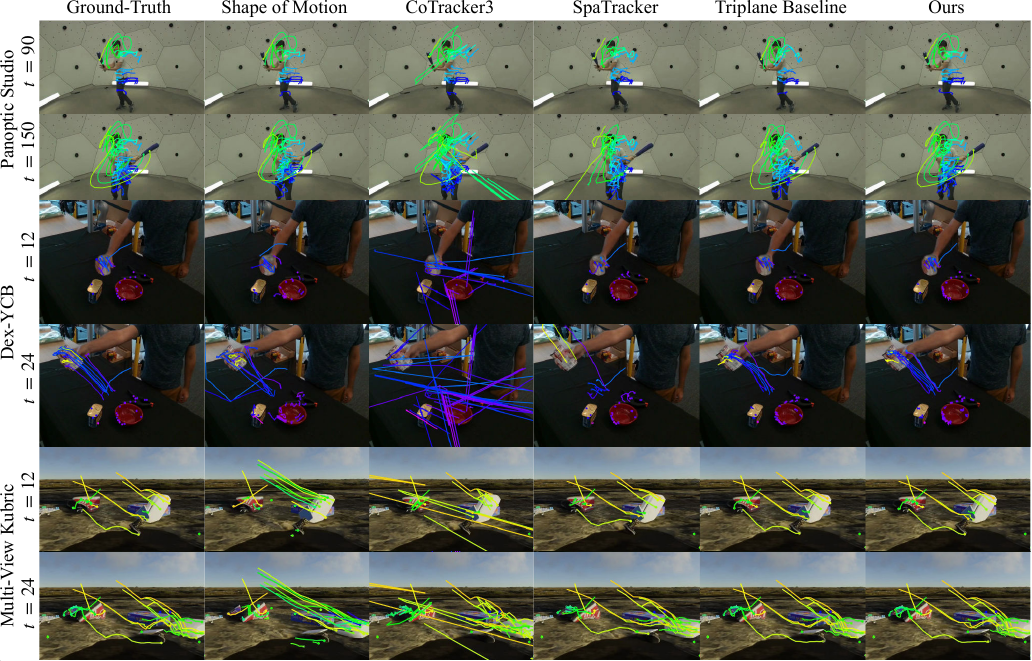}
    \vspace{-0.25in}
    \caption{
    \textbf{Qualitative comparison of multi-view 3D point tracking.} We visualize results from a camera viewpoint \emph{not} used during inference (and not near the input cameras). Each pair of rows corresponds to two time steps from the same dataset. The leftmost column shows ground-truth 3D trajectories, while remaining columns depict predictions from different methods. Points predicted as occluded are indicated by empty circles. Compared to baselines, our method more accurately maintains correspondences across views and handles occlusions. Please see the supp. video and project page for additional visualizations. These examples correspond to the results in \cref{tab:main-results-table}.
    }
    \label{fig:qualitative}
    \vspace{-0.9cm}
\end{figure*}

\cref{tab:main-results-table} compares MVTracker against several baselines on Panoptic Studio, DexYCB, and Multi-View Kubric. Our method consistently achieves higher location accuracy (\davg), better occlusion-aware tracking (AJ), and lower trajectory errors (MTE). This is visually apparent in \cref{fig:qualitative}. 

On Panoptic Studio, our tracker obtains an AJ of \B{86.0} and a \davg of \B{94.7}, substantially outperforming single-view methods such as SpaTracker (61.5\,AJ) and CoTracker3 (74.5\,AJ). We observe similar trends on DexYCB, where our method reaches an AJ of \B{71.6} with a median trajectory error of only \B{2.0}\,cm, compared to the lower performance of LocoTrack, DELTA, and CoTracker3. This is because 2D point trackers are not robust to lifting their 2D trajectories into 3D estimates based on estimated depth maps. Finally, on the Multi-View Kubric validation set, our model attains an AJ of \B{81.4} and an exceptionally low MTE of \B{0.7}\,cm.

Compared to optimization-based methods, which require per-sequence training, our approach is fully feed-forward and runs at 7.2 FPS. While Shape of Motion achieves reasonable accuracy, its iterative optimization makes it impractical for large-scale or real-time applications. Dynamic 3DGS further requires dense camera setups, limiting its applicability to real-world scenarios with fewer cameras.

\paragraph{Impact of the kNN-based correlation.}
Our method’s strong performance is largely due to its kNN-based correlation mechanism. Replacing it with a world-aligned triplane correlation leads to a significant drop in performance (see \cref{tab:main-results-table}). Triplane-based methods compress multi-view features onto fixed 2D planes, causing destructive feature collisions when different 3D points from the same or different views project to the same grid cell. This leads to information loss that cannot easily be disentangled, especially as the number of views increases. Moreover, triplanes require placing the 2D planes in a fixed scene-aligned coordinate frame with a pre-defined scale and extent, which is difficult to generalize across diverse camera setups and scene scales. In contrast, our kNN-based correlation operates directly in 3D world space and dynamically selects relevant neighbors, avoiding these issues and enabling more robust tracking.

\subsection{Ablations} \label{sec:ablations} 
We further evaluate the impact of the point correlation components, the varying number of input views, and different camera setups, as well as training augmentation strategies.

\begin{table}
\setlength{\tabcolsep}{2pt}
\centering
\caption{\textbf{Point Correlation Components.}
``No offset'' omits the offset vector entirely, 
``Offset + location'' concatenates both the offset and the neighbor's world-space coordinates, 
and ``Offset only'' encodes just the relative direction. 
Including only the offset vector (third row) yields the best overall performance. 
These experiments were trained for 25\% of the total steps and on 8 GPUs. 
}
\vspace{-0.3cm}
\begin{tabular}{l cccc}
                  \toprule
                  & \multicolumn{4}{c}{Multi-View Kubric~\cite{greff2022kubric}} \\
                  \midrule
                                    &   AJ \ua & \davg\ua &   OA \ua &  MTE \da \\
\midrule
No offset         &    21.3  &    45.3  &    40.6  &    15.6  \\
Offset + location &    48.7  &    59.6  &    68.5  &     6.8  \\\rowcolor{cyan!16}
Offset only       & \B{53.6} & \B{64.9} & \B{73.4} &  \B{4.3} \\
\bottomrule
\end{tabular}
\label{tab:ablate-model-correlation}
\end{table}

\paragraph{Point Correlation Components.}
\cref{tab:ablate-model-correlation} analyzes the impact of different components in our correlation module. Unlike 2D correlation, where relative direction is implicitly encoded in the pixel grid, kNN in 3D retrieves neighbors from all directions, making explicit offset information crucial. We compare three variants: (i) no offset vector, (ii) offset vector plus explicit neighbor location, and (iii) offset vector only. We train each variant for 25\% of total steps on 8 GPUs. The results suggest that (iii) “offset only” achieves the best performance across metrics, indicating that encoding only the relative offset, without concatenating absolute neighbor locations, yields most effective correspondences.

\begin{figure}
    \centering
    \includegraphics[width=\linewidth]{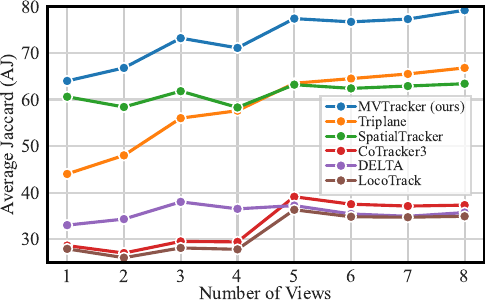}
    \vspace{-0.27in}
    \caption{\textbf{Effect of the Number of Input Views} on DexYCB~\cite{dexycb} (DUSt3R-based depth). MVTracker (blue) consistently improves with more views, reaching an AJ of 79.2 with eight views, indicating the benefit of multi-view information. SpatialTracker (green) and Triplane (orange) show moderate improvements but plateau earlier, while single-view methods such as CoTracker3, DELTA, and LocoTrack (red, purple, and brown) exhibit limited gains.
    }
    \label{fig:ablate-number-of-views}
\end{figure}

\paragraph{Effect of the Number of Input Views.}  
\cref{fig:ablate-number-of-views} shows how performance changes with the number of input views on DexYCB. Our tracker consistently improves as more views are added, achieving an AJ of 64.0 with a single view, 71.1 with four views, and 79.2 with eight views. This trend highlights the importance of multi-view information in reducing depth ambiguities and improving spatial consistency.

Compared to baseline methods, our approach benefits the most from additional views, thus suggesting better scalability. While SpatialTracker and Triplane show moderate improvements with more views, their performance plateaus earlier, indicating difficulty in fully leveraging additional multi-view information. In contrast, single-view methods like CoTracker3, DELTA, and LocoTrack show limited gains, reinforcing their inability to reliably reconstruct 3D correspondences. Similar trends are observed on Panoptic Studio as well as Multi-View Kubric (see \cref{sec:number-of-views-details}).

\begin{table}
\setlength{\tabcolsep}{2pt}
\centering
\caption{
    \textbf{Impact of Camera Setups.}
    Evaluation of AJ under varied camera setups. For Panoptic Studio~\cite{koppula2024tapvid}, we select 4 out of the available 27 views as opposing views (setup A) or nearby views (setups B and C). For DexYCB~\cite{dexycb}, we form 4-view sets out of the available 8 views: (A) views 1--4, (B) views 3--6, (C) views 5--8.
}
\vspace{-0.3cm}
\begin{tabular}{l ccc c ccc}
\toprule
  \multirow{2}{*}[-0.15em]{Method}
& \multicolumn{3}{c}{PStudio~\cite{koppula2024tapvid}}
& \xspace
& \multicolumn{3}{c}{DexYCB~\cite{dexycb}}
\\ \cmidrule(lr){2-4} \cmidrule(lr){6-8}
                                             &       A  &       B  &       C  &&       A  &       B  &       C  \\ \hline
Dynamic 3DGS~\cite{luiten2023dynamic}        &    66.5  &    50.8  &    56.6  &&    45.7  &      --  &      --  \\
Shape of Motion~\cite{som2024}               &    72.6  &    64.3  &    66.8  &&    36.2  &      --  &      --  \\
LocoTrack~\cite{cho2024local}                &    65.8  &    57.9  &    63.7  &&    27.8  &    40.9  &    42.9  \\
DELTA~\cite{ngo2024delta}                    &    68.1  &    61.1  &    65.9  &&    36.5  &    43.3  &    47.6  \\
CoTracker2~\cite{karaev2023cotracker}        &    69.5  &    62.3  &    66.4  &&    28.8  &    42.0  &    44.4  \\
CoTracker3~\cite{karaev24cotracker3}         &    74.5  &    66.3  &    70.9  &&    29.4  &    43.8  &    46.3  \\
SpaTracker~\cite{xiao2024spatialtracker}     &    61.5  &    54.8  &    57.8  &&    58.3  &    57.9  &    63.8  \\
Triplane Baseline                            &    65.1  &    59.9  &    63.5  &&    57.5  &    62.0  &    66.3  \\
\rowcolor{cyan!16} MVTracker (ours)          & \B{86.0} & \B{75.7} & \B{83.2} && \B{71.0} & \B{71.2} & \B{78.3} \\
\bottomrule
\end{tabular}
\label{tab:ablate-camera-setups}
\end{table}

\paragraph{Impact of Camera Setups.}  
\cref{tab:ablate-camera-setups} shows the performance (measured in AJ) under different camera configurations on Panoptic Studio and DexYCB. For Panoptic Studio, we selected sets of four cameras from the available 27, ensuring they were either positioned opposite each other (Setup A) or placed nearby with varying baseline sizes (Setups B and C). For DexYCB, we sequentially chose four out of the available eight cameras: views 1--4 for setup A, views 3--6 for setup B, and views 5--8 for setup C. Note that on Multi-View Kubric the camera configuration is randomized (with cameras generally oriented toward the scene center), so separate evaluations are unnecessary; the main table (\cref{tab:main-results-table}) already reflects a random sampling of camera placements. Our method consistently achieves high AJ across all tested setups, indicating robustness to varying camera positions.

\paragraph{Ablation of Training Augmentation.}
In \cref{app:sec:trainingdetails}, we examine how training augmentations impact performance, focusing on (i) the number of views used during training (ranging from 1 to 8) and (ii) the depth map source (ground truth vs. off-the-shelf estimation). Our results show that training with a variable number of views significantly improves robustness to different camera setups, ensuring better generalization across datasets. Additionally, incorporating both accurate and estimated depth maps helps mitigate domain shifts and enhances adaptability to real-world depth estimation errors. This is particularly noticeable on DexYCB, where DUSt3R depth quality varies significantly.

Overall, these ablations suggest that MVTracker remains robust across different camera configurations, benefits from diverse training conditions, and performs best with kNN-based offset encoding and carefully designed augmentations. Please refer to \cref{app:sub:ablation} for more ablation experiments and additional discussions. The inference speed measurements in \cref{app:sec:inference} show that MVTracker runs at 7.2 FPS when provided with sensor depth, underscoring its suitability for near-real-time and large-scale applications.

\section{Discussion and Future Work} \label{sec:discussion}
While our method effectively tracks 3D points across multiple views, it has limitations. The key challenge is its reliance on the quality of sparse-view depth estimation. If no sensor depth is available, the approach assumes a reasonably accurate estimated depth. However, in sparse camera setups, such estimation can be unreliable or fail entirely, making tracking infeasible. While our learned motion priors help mitigate moderate noise or incompleteness, a more principled solution would involve jointly estimating depth and tracking for potential mutual refinement, or developing a foundation model for 4D reconstruction with integrated tracking capabilities. We view this as a critical future direction for the community. \cref{app:depth-analysis} provides additional analysis on the impact of depth quality, robustness to noise, comparisons across different estimators, and failure cases.

Another limitation is that our study focuses on tracking within bounded regions where camera overlap is sufficient. Extending to unbounded or outdoor environments presents additional challenges due to limited training data availability, scene scale variation, and less constrained viewpoints.

A related issue is scene normalization: our model is trained on a fixed dataset with randomized but similar scene scales and layouts. To bridge distribution gaps at test time, we currently apply manually or heuristically determined similarity transforms. While this works well on our benchmarks and a few additional qualitative datasets, more principled approaches are needed to support arbitrary new scenes.

Finally, long-term 3D point tracking faces the broader challenge of scarce large-scale real-world training data. Unlike optimization-based approaches, data-driven point trackers require diverse training distributions to generalize. However, most existing datasets are synthetic, limiting robustness. Recent community efforts~\cite{karaev24cotracker3,doersch2024bootstap} show promise in leveraging self-supervised learning from real-world videos, which could be instrumental for improving generalization.

\vfill
\section{Conclusion}
\label{sec:conclusion}
We introduce MVTracker, the first \emph{data-driven multi-view 3D point tracker}, which combines kNN-based correlation in a fused 3D point cloud with a spatiotemporal transformer to track arbitrary points across multiple views. Our method outperforms single-view, multi-view, and optimization-based baselines, suggesting strong generalization across diverse camera configurations and occluded environments.

We hope that our work enables real-world use cases in robotics~\cite{10611409}, provides a better data-driven multi-view prior for optimization-based reconstruction methods~\cite{som2024,lei2024mosca,luiten2023dynamic}, and acts as a point tracking head in the upcoming age of scalable end-to-end 4D reconstruction and tracking models.

\vfill
\section*{Acknowledgements}
\begin{fulljustify}
    We are grateful for insightful discussions with Yiming Wang, Zador Pataki, Luigi Piccinelli, Paul‑Edouard Sarlin, and Philipp Lindenberger. We thank Ignacio Rocco and Skanda Koppula for clarifications regarding TAPVid-3D. This study was conducted within the national “Proficiency” research project (No. PFFS-21-19) funded by the Swiss Innovation Agency Innosuisse in 2021 as one of 15 flagship initiatives. This work was additionally supported as part of the Swiss AI initiative by a grant from the Swiss National Supercomputing Centre (CSCS) under project IDs A03 and A136 on Alps, enabling large-scale training and evaluation.
\end{fulljustify}

\clearpage
{
    \small
    \bibliographystyle{ieeenat_fullname}
    \bibliography{main}
}

\clearpage
\maketitlesupplementary
\appendix
\setcounter{page}{1}
\setcounter{table}{0}
\setcounter{figure}{0}
\setcounter{equation}{0}
\counterwithin{figure}{section}
\counterwithin{table}{section}
\renewcommand{\thetable}{\thesection.\arabic{table}}
\renewcommand{\thefigure}{\thesection.\arabic{figure}}
\renewcommand{\theequation}{\thesection.\arabic{equation}}

\section{Depth Estimation Analysis}
\label{app:depth-analysis}
\begin{figure}[h]
    \centering
    \includegraphics[width=\linewidth]{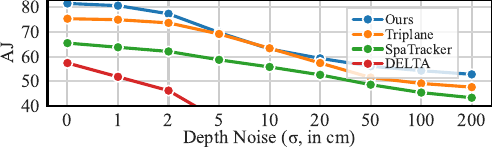}
    \vspace{-0.7cm}
    \caption{
    \textbf{Robustness to depth noise $\mathcal{N}(0, \sigma^2)$ on MV-Kubric.}}
    \label{fig:depth-noise-ablation-v4}
\end{figure}

\paragraph{Robustness to depth quality.}
Our method is designed to be tolerant of moderate noise in estimated depth. As shown in \cref{fig:depth-noise-ablation-v4}, tracking performance remains stable under additive Gaussian noise up to $\sigma$ = 2\,cm. This robustness stems from extensive training augmentations and from the transformer’s ability to interpolate and re-identify points even when partial depth information is missing or corrupted. 

\paragraph{Depth source comparisons.}
\cref{tab:depth-types} summarizes performance across different depth sources, including sensor depth, optimization-based reconstruction, and learned estimators (DUSt3R and VGGT). Our method achieves strong performance with noisy but plausible depth maps from learned estimators, particularly VGGT, but degrades significantly when depth estimation produces misaligned geometry or fails.

\paragraph{DUSt3R alignment and runtime.}
To align DUSt3R~\cite{wang2024dust3r} reconstructions to our camera setup, we globally optimize the pairwise maps using its built-in alignment (following Sec.~3.4 in DUSt3R), while keeping our camera intrinsics and extrinsics fixed. This reconstruction step takes 300 iterations and runs at 0.17 FPS. VGGT offers a significantly faster alternative at 3.1 FPS.

\begin{figure}[h]
    \centering  
    \includegraphics[width=\linewidth]{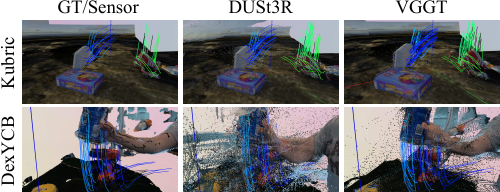}
    \vspace{-0.7cm}
    \caption{\textbf{Different depth sources along our predicted tracks.}}
    \label{fig:depths}
\end{figure}

\paragraph{Discussion.}
Sparse-view reconstruction remains an open problem. In our evaluations, DUSt3R was the most reliable method for producing usable depth maps, but even it fails on some scenes. As depth estimation continues to improve, particularly with recent advances in foundation models for geometry (e.g., MegaSAM~\cite{li2024_megasam}, Easi3R~\cite{chen2025easi3r}), MVTracker can directly benefit (with or without retraining). \cref{fig:depths} visualizes our predicted tracks under different depth sources, and \cref{tab:depth-types} quantifies performance across depth sources.

\begin{table}[h]
\centering
\caption{\textbf{AJ for different depth sources.} GT*: optimization-based depth. D: DUSt3R. V: VGGT. S: Sensor. \ding{55}: depth estimation failed.}
\vspace{-0.35cm}
\resizebox{\columnwidth}{!}{%
\begin{tabular}{l ccc ccc ccc}
\toprule
\multirow{2}{*}{Method}
& \multicolumn{3}{c}{Panoptic Studio}
& \multicolumn{3}{c}{DexYCB}
& \multicolumn{3}{c}{MV-Kubric}
\\
\cmidrule(lr){2-4} \cmidrule(lr){5-7} \cmidrule(lr){8-10}
& GT* & D & V & S & D & V & GT & D & V \\
\midrule
DELTA              &    68.1  & \ding{55} &     1.8  &    54.6  &    36.8  &    21.1  &    57.4  &    25.7  &     1.9  \\
SpaTracker         &    61.5  & \ding{55} &    10.6  &    60.9  &    58.3  &    33.2  &    65.5  &    61.7  &     3.9  \\
Triplane Baseline  &    65.1  & \ding{55} &    29.7  &    68.0  &    57.5  &    61.8  &    74.7  & \B{70.7} &    46.7  \\ \rowcolor{cyan!16}
MVTracker (ours)   & \B{86.0} & \ding{55} & \B{47.7} & \B{82.7} & \B{71.6} & \B{68.0} & \B{81.4} &    70.0  & \B{48.7} \\
\bottomrule
\end{tabular}
}
\label{tab:depth-types}
\end{table}

\section{Inference Speed Comparison}
\label{app:sec:inference}

We evaluate the runtime of various methods by measuring frames per second (FPS) on a representative test configuration. \cref{tab:fps_comparison} summarizes the results, excluding depth-estimation time. For \emph{optimization-based} methods such as Shape of Motion~\cite{som2024} and Dynamic 3D Gaussians~\cite{luiten2023dynamic}, the per-sequence runtime is significantly higher (\eg, about 30 minutes for Shape of Motion and about 50 minutes for Dynamic 3DGS on a single Panoptic Studio~\cite{koppula2024tapvid} sequence). Overall, the optimization-based approaches are orders of magnitude slower, making them less practical for large-scale applications and unusable for real-time applications. In contrast, our feed-forward model runs online at a comparable speed to the triplane-based baseline while outperforming it in accuracy, highlighting the efficiency of our kNN correlation in 3D point clouds. Note that we correctly adjust the effective FPS by taking into account the window size for sliding-window-based methods.

\begin{table}[h]
\setlength{\tabcolsep}{2pt}
\centering
\caption{\textbf{Runtime Comparison (FPS).} We measure frames per second for each method under a representative test setting, excluding depth-estimation overhead. Optimization-based methods are indicated with placeholders for their much lower throughput.}
\vspace{-0.3cm}
\begin{tabular}{l c}
\toprule
\textbf{Method} & \textbf{FPS} \\
\midrule
Dynamic 3DGS~\cite{luiten2023dynamic}     & \emph{(optimization-based)} \\
Shape of Motion~\cite{som2024}            & \emph{(optimization-based)} \\
SpaTracker~\cite{xiao2024spatialtracker} &  1.4 \\
DELTA~\cite{ngo2024delta}                &  1.5 \\
SceneTracker~\cite{wang2024scenetracker} &  3.5 \\ %
CoTracker2~\cite{karaev2023cotracker}    &  5.7 \\
Triplane Baseline                        &  5.8 \\
\rowcolor{cyan!16}MVTracker (ours)      &  7.2 \\ %
CoTracker3~\cite{karaev24cotracker3}     & 18.4\\
\bottomrule
\end{tabular}
\label{tab:fps_comparison}
\end{table}

\section{Ablation Study} \label{app:sub:ablation}

\subsection{Impact of Different Depth Sources}
\Cref{tab:dexycb-depth-types} reports the performance of our model and several baselines on DexYCB when using two different depth sources: DUSt3R~\cite{wang2024dust3r} (estimated) and sensor-based depth. We observe that sensor-based depth leads to improved accuracy for all methods, highlighting the importance of depth quality. Notably, our multi-view tracker benefits substantially from sensor depth, surpassing single-view methods and the triplane baseline by a larger margin. This suggests that our fused 3D point cloud representation is robust to a wide range of depth sources, yet can further leverage higher-fidelity depth to achieve stronger 3D point tracking.
\begin{table}
\setlength{\tabcolsep}{2pt}
\centering
\caption{\textbf{Impact of Different Depth Sources} on DexYCB. We report results using either DUSt3R-estimated~\cite{wang2024dust3r} or sensor-based depth for all methods. Sensor depth generally improves performance across the board, underscoring the importance of high-fidelity depth. Notably, our multi-view tracker benefits the most, indicating that our fused 3D point cloud representation effectively capitalizes on better depth quality.
}
\vspace{-0.3cm}
\begin{tabular}{l cccc}
                  \toprule
                                    Method &   AJ\hspace{-0.01cm}\ua & \davg\hspace{-0.05cm}\ua &   OA\hspace{-0.03cm}\ua\hspace{0.03cm} &  \hspace{-0.1cm}MTE\hspace{-0.03cm}\da \\
                  \midrule
                \multicolumn{5}{r}{DexYCB~\cite{dexycb} (DUSt3R depth) } \\
\midrule
LocoTrack~\cite{cho2024local}                &    27.8  &    38.6  &    77.0  &    22.8  \\ 
DELTA~\cite{ngo2024delta}                    &    36.8  &    51.6  &    61.0  &    18.3  \\ 
CoTracker2~\cite{karaev2023cotracker}        &    28.8  &    40.5  &    76.2  &    20.8  \\ 
CoTracker3~\cite{karaev24cotracker3}         &    29.4  &    40.3  &    78.6  &    22.0  \\ 
SpaTracker~\cite{xiao2024spatialtracker}     &    58.3  &    72.0  &    80.2  &     5.9  \\ 
SpaTrackerV2~\cite{xiao2025spatialtrackerv2} &    35.5  &    45.1  &    91.3  &     9.8  \\
TAPIP3D~\cite{tapip3d}                       &    38.8  &    50.0  &    90.1  &     8.2  \\
Triplane Baseline                            &    57.5  &    71.0  &    81.3  &     4.3  \\ 
\rowcolor{cyan!16} MVTracker (ours)          & \B{71.6} & \B{80.6} & \B{91.3} &  \B{2.0} \\

                  \toprule
                 \multicolumn{5}{r}{DexYCB~\cite{dexycb} (sensor depth)} \\
                  \toprule
LocoTrack~\cite{cho2024local}                &    55.1  &    64.6  &    76.7  &    27.5  \\
DELTA~\cite{ngo2024delta}                    &    54.6  &    67.8  &    72.6  &    42.6  \\
CoTracker2~\cite{karaev2023cotracker}        &    56.4  &    67.4  &    76.4  &    26.4  \\
CoTracker3~\cite{karaev24cotracker3}         &    57.3  &    68.5  &    76.7  &    23.4  \\
SpaTracker~\cite{xiao2024spatialtracker}     &    60.9  &    75.0  &    82.5  &     4.0  \\
SpaTrackerV2~\cite{xiao2025spatialtrackerv2} &    69.2  &    77.5  &    91.3  &     4.2  \\
TAPIP3D~\cite{tapip3d}                       &    77.2  &    88.9  &    88.0  &     1.3  \\
Triplane Baseline                            &    68.0  &    79.5  &    85.9  &     2.6  \\
\rowcolor{cyan!16} MVTracker (ours)          & \B{82.7} & \B{91.0} & \B{91.6} &  \B{0.8} \\
\bottomrule
\end{tabular}
\label{tab:dexycb-depth-types}
\end{table}

\subsection{Varying the Number of Input Views}
\label{sec:number-of-views-details}
\cref{tab:ablate-number-of-views} shows detailed per-dataset results for a varied number of input views. Our tracker consistently improves as more views are added.

\begin{table}[t]
\setlength{\tabcolsep}{1.2pt}
\centering
\caption{\textbf{Varying the Number of Input Views.} Evaluation of AJ under a varied number of input views.}
\vspace{-0.3cm}
\begin{tabular}{lcccccccc}
                                             \toprule \multirow{2}{*}[0.07em]{Method} & \multicolumn{8}{c}{Number of views} \\
                                             &       1  &       2  &       3  &        4 &       5  &       6  &       7  &       8  \\
                                             \toprule
                                             & \multicolumn{8}{c}{Panoptic Studio~\cite{koppula2024tapvid}} \\
                                             \toprule
LocoTrack~\cite{cho2024local}                &     54.7 &    56.4  &    61.4  &    65.8  &    65.8  &    67.3  &    66.4  &    66.8  \\
DELTA~\cite{ngo2024delta}                    &     60.9 &    61.8  &    63.7  &    68.1  &    66.7  &    67.4  &    67.6  &    66.3  \\
CoTracker2~\cite{karaev2023cotracker}        &     59.5 &    60.7  &    64.4  &    69.5  &    68.8  &    69.3  &    68.4  &    69.9  \\
CoTracker3~\cite{karaev24cotracker3}         &     61.9 &    65.0  &    68.5  &    74.5  &    74.0  &    75.2  &    75.1  &    76.0  \\
SpaTracker~\cite{xiao2024spatialtracker}     &     51.2 &    52.1  &    57.5  &    61.5  &    60.9  &    61.4  &    62.6  &    63.4  \\
SpaTrackerV2~\cite{xiao2025spatialtrackerv2} &     57.4 &    62.7  &    66.8  &    72.4  &    71.8  &    72.7  &    72.7  &    72.9  \\
TAPIP3D~\cite{tapip3d}                       &  \B{68.8}& \B{72.9} &    78.5  &    84.3  &    84.6  &    85.6  &    86.0  &    86.8  \\
Triplane Baseline                            &     61.9 &    60.9  &    61.6  &    65.0  &    64.0  &    65.1  &    66.8  &    65.6  \\
\rowcolor{cyan!16} MVTracker (ours)          &    66.4  &    72.2  & \B{79.0} & \B{86.0} & \B{87.4} & \B{88.8} & \B{89.1} & \B{89.9} \\
                                             \toprule
                                             & \multicolumn{8}{c}{DexYCB~\cite{dexycb} (DUSt3R depth)} \\
                                             \toprule
LocoTrack~\cite{cho2024local}                &     27.9 &    26.0  &    28.1  &    27.8  &    36.3  &    34.8  &    34.7  &    34.9  \\
DELTA~\cite{ngo2024delta}                    &     33.0 &    34.3  &    38.0  &    36.5  &    37.2  &    35.4  &    34.9  &    35.7  \\
CoTracker2~\cite{karaev2023cotracker}        &     29.8 &    26.4  &    29.2  &    28.8  &    37.8  &    36.2  &    36.0  &    36.0  \\
CoTracker3~\cite{karaev24cotracker3}         &     28.6 &    27.0  &    29.5  &    29.4  &    39.1  &    37.5  &    37.1  &    37.3  \\
SpaTracker~\cite{xiao2024spatialtracker}     &     60.6 &    58.4  &    61.8  &    58.3  &    63.2  &    62.4  &    62.9  &    63.4  \\
SpaTrackerV2~\cite{xiao2025spatialtrackerv2} &     39.8 &    39.5  &    36.5  &    35.5  &    41.1  &    37.1  &    37.0  &    37.7  \\
TAPIP3D~\cite{tapip3d}                       &     36.6 &    35.6  &    40.5  &    38.8  &    57.7  &    54.2  &    55.2  &    56.4  \\
Triplane Baseline                            &     44.0 &    48.0  &    56.0  &    57.6  &    63.5  &    64.5  &    65.5  &    66.8  \\
\rowcolor{cyan!16} MVTracker (ours)          & \B{64.0} & \B{66.8} & \B{73.2} & \B{71.1} & \B{77.4} & \B{76.7} & \B{77.3} & \B{79.2} \\
                                             \toprule
                                             & \multicolumn{8}{c}{Multi-View Kubric~\cite{greff2022kubric}} \\
                                             \toprule
LocoTrack~\cite{cho2024local}                &    57.3  &    56.1  &    53.8  &    52.5  &    52.3  &    52.1  &    51.6  &    51.8  \\
DELTA~\cite{ngo2024delta}                    &    65.9  &    61.9  &    58.9  &    57.4  &    56.3  &    56.0  &    54.8  &    54.1  \\
CoTracker2~\cite{karaev2023cotracker}        &    61.4  &    59.1  &    56.1  &    54.6  &    53.8  &    53.7  &    53.1  &    52.8  \\
CoTracker3~\cite{karaev24cotracker3}         &    60.1  &    58.7  &    56.5  &    55.1  &    54.8  &    54.6  &    54.2  &    54.3  \\
SpaTracker~\cite{xiao2024spatialtracker}     &    72.8  &    70.3  &    67.4  &    65.5  &    64.8  &    64.5  &    63.1  &    62.7  \\
SpaTrackerV2~\cite{xiao2025spatialtrackerv2} &     54.0 &    57.6  &    57.6  &    58.6  &    58.3  &    59.1  &    58.8  &    59.7  \\
TAPIP3D~\cite{tapip3d}                       &     76.4 &    75.4  &    72.8  &    72.5  &    71.7  &    71.4  &    70.9  &    70.6  \\
Triplane Baseline                            &    71.7  &    73.4  &    74.1  &    74.7  &    75.0  &    75.2  &    75.2  &    75.7  \\
\rowcolor{cyan!16} MVTracker (ours)          & \B{81.7} & \B{81.2} & \B{81.4} & \B{81.4} & \B{81.9} & \B{82.1} & \B{82.3} & \B{82.5} \\
\bottomrule
\end{tabular}
\label{tab:ablate-number-of-views}
\end{table}

\subsection{2D Tracking Accuracy} 
\begin{table}
\setlength{\tabcolsep}{2pt}
\centering
\caption{\textbf{2D Tracking Evaluation.} Evaluation of projected 2D tracking performance using standard 2D metrics~\cite{doersch2023tapvid}. Multi-view outputs are projected onto four views and the metrics are averaged across these views. Note that we can only report location accuracies since our multi-view methods report any-view visibility (not per-view).}
\vspace{-0.3cm}
\begin{tabular}{lcccccccc}
                                             \toprule {Method}
                                             &       $\delta_\text{avg}^{\text{2D}}$\ua
                                             &       $\delta_{< 1}^{\text{2D}}$\ua
                                             &       $\delta_{< 2}^{\text{2D}}$\ua
                                             &       $\delta_{< 4}^{\text{2D}}$\ua
                                             &       $\delta_{< 8}^{\text{2D}}$\ua
                                             &       $\delta_{<16}^{\text{2D}}$\ua
                                             \vspace{0.05cm}
                                             \\
                                             \toprule
                                             & \multicolumn{6}{c}{Panoptic Studio~\cite{koppula2024tapvid}} \\
                                             \toprule
LocoTrack~\cite{cho2024local}                &    66.7  &    29.3  &    51.4  &    72.6  &    86.5  &    93.7  \\
DELTA~\cite{ngo2024delta}                    &    70.0  & \B{34.8} &    56.5  &    76.5  &    87.6  &    94.8  \\
CoTracker2~\cite{karaev2023cotracker}        &    64.3  &    28.0  &    49.3  &    69.7  &    82.9  &    91.4  \\
CoTracker3~\cite{karaev24cotracker3}         &    70.9  &    33.1  & \B{56.8} &    78.2  &    90.5  &    96.2  \\
SpaTracker~\cite{xiao2024spatialtracker}     &    66.4  &    29.3  &    51.8  &    71.8  &    85.7  &    93.3  \\
Triplane Baseline                            &    51.9  &     8.5  &    24.1  &    52.0  &    80.3  &    94.5  \\
\rowcolor{cyan!16} MVTracker (ours)         & \B{70.5} &    26.6  &    52.9  & \B{79.5} & \B{94.6} & \B{98.9} \\
                                             \toprule
                                             & \multicolumn{6}{c}{DexYCB~\cite{dexycb} (sensor depth)} \\
                                             \toprule
LocoTrack~\cite{cho2024local}                &    83.7  &    74.1  &    78.4  &    84.0  &    89.4  &    92.4  \\
DELTA~\cite{ngo2024delta}                    &    84.6  &    75.7  &    79.9  &    85.2  &    89.4  &    92.8  \\
CoTracker2~\cite{karaev2023cotracker}        &    83.7  &    73.7  &    78.9  &    84.5  &    89.3  &    92.1  \\
CoTracker3~\cite{karaev24cotracker3}         &    84.1  &    74.8  &    79.0  &    84.5  &    89.4  &    92.9  \\
SpaTracker~\cite{xiao2024spatialtracker}     &    85.8  & \B{75.9} &    80.7  &    86.1  &    91.2  &    94.9  \\
Triplane Baseline                            &    75.9  &    52.2  &    70.4  &    78.9  &    85.9  &    92.0  \\
\rowcolor{cyan!16} MVTracker (ours)         & \B{87.5} &    73.3  & \B{81.2} & \B{88.3} & \B{95.5} & \B{99.2} \\
                                             \toprule
                                             & \multicolumn{6}{c}{Multi-View Kubric~\cite{greff2022kubric}} \\
                                             \toprule
LocoTrack~\cite{cho2024local}                &    75.3  &    53.6  &    66.8  &    78.0  &    86.1  &    91.9  \\
DELTA~\cite{ngo2024delta}                    &    78.4  &    61.9  &    72.0  &    80.3  &    86.5  &    91.2  \\
CoTracker2~\cite{karaev2023cotracker}        &    72.6  &    51.8  &    63.7  &    74.2  &    83.1  &    90.2  \\
CoTracker3~\cite{karaev24cotracker3}         &    74.8  &    54.2  &    66.3  &    76.8  &    85.4  &    91.5  \\
SpaTracker~\cite{xiao2024spatialtracker}     &    75.8  &    51.4  &    65.8  &    79.3  &    88.5  &    94.0  \\
Triplane Baseline                            &    80.6  &    51.3  &    72.4  &    87.3  &    94.5  &    97.7  \\
\rowcolor{cyan!16} MVTracker (ours)         & \B{86.8} & \B{61.2} & \B{81.7} & \B{94.0} & \B{97.9} & \B{99.2} \\
\bottomrule
\end{tabular}
\label{tab:ablate-per-view-2d-point-tracking-performance}
\end{table}

\cref{tab:ablate-per-view-2d-point-tracking-performance} compares the performance of 2D point tracking methods when their outputs are projected into 2D using standard metrics~\cite{doersch2023tapvid}. Although 2D methods do not rely on estimated depth quality, they generally achieve lower location accuracies compared to our multi-view approach, which fuses multi-view data to enhance performance. On DexYCB, we only report results using sensor depth maps to provide a more informative comparison in 2D point tracking. For results with depth estimated by DUSt3R, where single-view methods benefit from not having to handle noisy or incomplete depth maps, see \cref{tab:ablate-per-view-2d-point-tracking-performance-dexycb}. Note that such depth noise or incompleteness affects our 2D tracking more than that of SpaTracker, due to SpaTracker’s XY-aligned feature plane compared to our fused 3D point cloud.

\begin{table}
\setlength{\tabcolsep}{2pt}
\centering
\caption{Evaluation of projected 2D tracking performance using standard 2D metrics~\cite{doersch2023tapvid} on DexYCB with estimated depths. Single-view methods only require RGB for tracking and do not depend on the quality of the depth for achieving high 2D point tracking performance. However, the multi-view methods face challenges when dealing with the noisy depth estimates on DexYCB.}
\vspace{-0.3cm}
\begin{tabular}{lcccccccc}
                                             \toprule {Method}
                                             &       $\delta_\text{avg}^{\text{2D}}$\ua
                                             &       $\delta_{< 1}^{\text{2D}}$\ua
                                             &       $\delta_{< 2}^{\text{2D}}$\ua
                                             &       $\delta_{< 4}^{\text{2D}}$\ua
                                             &       $\delta_{< 8}^{\text{2D}}$\ua
                                             &       $\delta_{<16}^{\text{2D}}$\ua
                                             \vspace{0.05cm}
                                             \\
                                             \toprule
                                             & \multicolumn{6}{c}{DexYCB~\cite{dexycb} (DUSt3R depth)} \\
                                             \toprule
LocoTrack~\cite{cho2024local}                &    85.4  &    75.6  &    80.0  &    85.8  &    91.4  &    94.4  \\
DELTA~\cite{ngo2024delta}                    &    85.6  & \B{77.0} &    81.2  &    86.0  &    90.2  &    93.7  \\
CoTracker2~\cite{karaev2023cotracker}        &    86.2  &    75.7  &    81.3  &    87.1  &    92.1  &    95.0  \\
CoTracker3~\cite{karaev24cotracker3}         & \B{86.7}  & \B{77.0} & \B{81.5} & \B{87.2} & \B{92.1} & \B{95.7} \\
SpaTracker~\cite{xiao2024spatialtracker}     &    86.0  &    76.4  &    80.8  &    86.2  &    91.7  &    95.1  \\
Triplane Baseline                            &    64.3  &    31.8  &    53.6  &    71.4  &    79.3  &    85.4  \\
\rowcolor{cyan!16} MVTracker (ours)          &    71.3  &    42.1  &    59.8  &    76.0  &    85.2  &    93.5  \\
\bottomrule
\end{tabular}
\label{tab:ablate-per-view-2d-point-tracking-performance-dexycb}
\end{table}

\subsection{Training Augmentation} \label{app:sec:trainingdetails}
\begin{table}
\setlength{\tabcolsep}{2pt}
\centering
\caption{\textbf{Training Augmentation} of the variable number of input views ranging from 1 to 8  (V) and varying depth sources between ground-truth and off-the-shelf depth estimation (D) during training on tracking performance.} 
\vspace{-0.3cm}
\begin{tabular}{cc c cccc}
\toprule
V & D                 &&   AJ \ua & \davg\ua &   OA \ua &  MTE \da \\
                      \toprule
                      &&& \multicolumn{4}{c}{Panoptic Studio~\cite{koppula2024tapvid}} \\
                      \toprule
\ding{51} &           &&    80.4  & \B{94.4} &    87.8  &  \B{3.1 }\\
          & \ding{51} && \B{80.7} &    93.3  & \B{88.7} &     3.5  \\\rowcolor{cyan!16}
\ding{51} & \ding{51} && \B{80.7} &    94.3  &    88.2  &     3.2  \\
                      \toprule
                      &&& \multicolumn{4}{c}{DexYCB~\cite{dexycb}} \\
                      \toprule
\ding{51} &           &&    49.9  &    72.4  &    70.5  &     3.6  \\
          & \ding{51} &&    62.3  &    77.1  &    82.1  &     3.2  \\\rowcolor{cyan!16}
\ding{51} & \ding{51} && \B{65.2} & \B{79.5} & \B{82.9} &  \B{2.3 }\\
                      \toprule
                      &&& \multicolumn{4}{c}{Multi-View Kubric~\cite{greff2022kubric}} \\
                      \toprule
\ding{51} &           &&    80.2  & \B{91.1}  &    91.1  &  \B{0.7 }\\
          & \ding{51} &&    77.2  &    88.0  &    91.0  &     0.8  \\\rowcolor{cyan!16}
\ding{51} & \ding{51} && \B{80.4} &    90.5  & \B{92.1} &  \B{0.7 }\\
\bottomrule
\end{tabular}
\label{tab:ablate-augmentation-for-main-4view-result}
\end{table}

\begin{table}
\setlength{\tabcolsep}{2pt}
\centering
\caption{\textbf{Training Augmentation} with evaluation on different numbers of views. We investigate the impact of using a variable number of views ranging from 1 to 8 (V) and varying depth sources between ground-truth and off-the-shelf depth estimation (D) during training on tracking performance. Results are reported in AJ.}
\vspace{-0.3cm}
\begin{tabular}{cccccccccc}
\toprule
\multirow{2}{*}[0.1em]{V} & \multirow{2}{*}[0.1em]{D} 
                      & \multicolumn{8}{c}{Number of views} \\
          &           &       1  &       2  &       3  &        4 &       5  &       6  &       7  &       8  \\
                      \toprule
                      && \multicolumn{8}{c}{Panoptic Studio~\cite{koppula2024tapvid}} \\
                      \toprule
\ding{51} &           & \B{69.3} &    72.0  &    76.9  &    80.4  &    81.4  & \B{84.4} &    84.8  &    85.1  \\
          & \ding{51} &    67.0  &    71.3  & \B{77.2} & \B{80.7} & \B{81.9} &    84.2  & \B{84.9} & \B{85.8} \\\rowcolor{cyan!16}
\ding{51} & \ding{51} &    68.1  & \B{73.1} &    76.9  & \B{80.7} &    81.5  &    83.4  &    84.4  &    84.3  \\
                      \toprule
                      && \multicolumn{8}{c}{DexYCB~\cite{dexycb} (DUSt3R depth)} \\
                      \toprule
\ding{51} &           &    37.6  &    40.0  &    48.1  &    49.9  &    65.6  &    67.4  &    69.6  &    73.4  \\
          & \ding{51} &    51.6  &    54.1  &    60.7  &    62.3  &    68.8  &    69.9  &    71.8  &    74.8  \\\rowcolor{cyan!16}
\ding{51} & \ding{51} & \B{56.3} & \B{58.9} & \B{64.9} & \B{65.2} & \B{74.0} & \B{75.4} & \B{76.3} & \B{78.9} \\
                      \toprule
                      && \multicolumn{8}{c}{Multi-View Kubric~\cite{greff2022kubric}} \\
                      \toprule
\ding{51} &           & \B{80.7} & \B{79.9} & \B{80.1} &    80.2  &    80.9  &    81.4  & \B{81.8} &    81.9  \\
          & \ding{51} &    75.9  &    76.3  &    76.6  &    77.2  &    77.2  &    78.0  &    78.1  &    78.4  \\\rowcolor{cyan!16}
\ding{51} & \ding{51} &    80.4  &    79.6  & \B{80.0} & \B{80.4} & \B{81.1} & \B{81.6} & \B{81.8} & \B{82.0} \\
\bottomrule
\end{tabular}
\label{tab:ablate-augmentation-for-varying-views}
\end{table}

\cref{tab:ablate-augmentation-for-main-4view-result} and~\ref{tab:ablate-augmentation-for-varying-views} evaluate the impact of training augmentation strategies. We consider two factors: (V) employing a variable number of views during training (ranging from 1 to 8) and (D) varying the source of depth maps (ground-truth versus off-the-shelf depth estimation). The results reveal that combining both variable view and depth augmentations leads to the best performance, especially as measured on DexYCB.

We apply a range of augmentations to improve generalization and robustness. These include photometric augmentations such as color jitter, Gaussian blur, and occlusion simulation via erasing and region replacement. Spatial augmentations involve random cropping, padding, flipping, and scaling. Depth perturbations include noise, rescaling, and occlusion-based erasure. Scene-level augmentations transform the world coordinate system through random rotation, translation, and scaling. Camera intrinsics and extrinsics are perturbed to simulate realistic variations. Additionally, we vary the number of tracks per sample, randomly sample the number of input views (between 1 and 6), and use either ground-truth depths or depths from an estimation method to improve model adaptability. 

\section{Baselines}
\label{app:sec:baseline_implementation}
\textbf{Dynamic 3D Gaussians.} We use the original training code provided by the authors of the Dynamic 3D Gaussians \cite{luiten2023dynamic}. The training process utilizes camera intrinsics, extrinsics, segmentation masks, images, and an initial point cloud. This initial point cloud is constructed from depth maps corresponding to selected viewpoints at the initial timestep. To differentiate between static and dynamic points within the point cloud, we leverage the provided segmentation masks.

For tracking, we identify the most influential Gaussian for a given point at time $t$ based on the influence computation described in the Dynamic 3DGS paper. To establish a trajectory, we then track the motion of this Gaussian across all time steps. The visibility of each point at time $t$ is determined by comparing the depth of the Gaussian with the corresponding depth value of the depth map rendered from each camera viewpoint. Trajectories that show sudden discontinuities or have a dominant Gaussian influence near zero for the query point are classified as static to ensure consistency in motion.

To evaluate the model's performance in novel view synthesis, we conducted experiments on Panoptic Studio. We trained different models using a varying number of cameras. Then, we render images using these trained models from four test cameras. Our results in \cref{tab:d3dgs-vary-number-of-views} show that with 27 input views, we achieve reconstruction quality comparable to that reported in the original paper. However, as the number of views decreases, reconstruction quality reduces.
\begin{table}[t]
\setlength{\tabcolsep}{4pt}
\centering
\caption{\textbf{Dynamic 3D Gaussians.} Novel view synthesis scores for Dynamic 3DGS~\cite{luiten2023dynamic} under varying numbers of cameras on Panoptic Studio~\cite{koppula2024tapvid}.}
\vspace{-0.3cm}
\begin{tabular}{lcccc}
\toprule
 & \multicolumn{4}{c}{Number of Cameras} \\
\cmidrule(lr){2-5}
 & 27 & 17 & 9 & 4 \\
\midrule
PSNR & 28.5 & 23.7 & 17.9 & 12.6 \\
SSIM & 0.91 & 0.83 & 0.74 & 0.52 \\
\bottomrule
\end{tabular}
\label{tab:d3dgs-vary-number-of-views}
\end{table}

\textbf{Shape of Motion.} We adapt the monocular training pipeline of the original Shape of Motion \cite{som2024} to a multi-view setting. The method requires segmentation masks for moving objects, depth maps, long-range 2D tracks, and camera poses. In our multi-view adaptation, we rely on the available depth maps, segmentation masks, and camera poses, while using TAPIR \cite{doersch2023tapir} to obtain long-range 2D tracks, as in the original approach.

For both static and dynamic part initialization, we use multi-view depth maps and segmentation masks to construct an initial 3D representation that captures the overall scene structure more effectively than the monocular approach. We follow the original paper's strategy for choosing the canonical frame and optimizing motion bases. Additionally, we adapt the monocular loss function to a multi-view formulation, computing the original loss function separately for each view and averaging the results across all views. This adjustment ensures that the optimization effectively integrates multi-view information.

\section{Evaluation on TAPVid-2D}
\label{app:tapvid2d}

In \cref{tab:tapvid2d}, we evaluate MVTracker on the TAPVid-2D benchmark by projecting our predicted 3D trajectories onto image views. As our method requires depth input, we report results using various monocular depth estimators. However, tracking performance is significantly degraded in this benchmark, which is in part attributable to TAPVid-2D containing outdoor unbounded scenes, our scene normalization not aligning scenes well to the training distribution, and monocular depth estimation frequently failing or flickering.

\begin{table*}
\setlength{\tabcolsep}{2pt}
\centering
\caption{\textbf{TAPVid-2D Evaluation.} Evaluation results in 2D point tracking on TAPVid-2D~\cite{doersch2023tapvid}. Since our method requires depth as input, we report results for several monocular depth estimators. However, performance is far from satisfactory for most trajectories, which is in part attributable to out-of-distribution outdoor scenes with many far-away points, scenes not being normalized well to the training scale, failures in depth estimation, flickering in the estimated video depth, etc. We report both the original TAPVid-2D metrics as well as the metrics used elsewhere in this paper; the difference is that our metrics take the average of per-trajectory metrics, whereas TAPVid-2D computes each metric over all points, treating all trajectories as one trajectory. All metrics are in pixel scale (\eg, $\delta_{< 4}^{\text{2D}}$ is the within-a-radius-of-4-pixels location accuracy, MTE is in pixels, etc.). Best and second-best results per metric are \B{bold} and \U{underlined}, respectively.}
\vspace{-0.2cm}
\begin{tabular}{ll c cccccc c cccccc c cc}
\toprule
Method
& Depth
& \hspace{1ex}
& {\scriptsize $\text{AJ}      ^{\text{2D}}$ }
& {\scriptsize $\text{AJ}_{< 1}^{\text{2D}}$ }
& {\scriptsize $\text{AJ}_{< 2}^{\text{2D}}$ }
& {\scriptsize $\text{AJ}_{< 4}^{\text{2D}}$ }
& {\scriptsize $\text{AJ}_{< 8}^{\text{2D}}$ }
& {\scriptsize $\text{AJ}_{<16}^{\text{2D}}$ }
& \hspace{1ex}
& {\scriptsize $\delta_\text{avg}^{\text{2D}}$ }
& {\scriptsize $\delta_{< 1}^{\text{2D}}$ }
& {\scriptsize $\delta_{< 2}^{\text{2D}}$ }
& {\scriptsize $\delta_{< 4}^{\text{2D}}$ }
& {\scriptsize $\delta_{< 8}^{\text{2D}}$ }
& {\scriptsize $\delta_{<16}^{\text{2D}}$ }
& \hspace{1ex}
& {\scriptsize OA }
& {\scriptsize MTE }
\\
\midrule
&&&\multicolumn{16}{c}{TAPVid-2D Metrics} \\
\midrule \rowcolor{cyan!16}
MVTracker (ours) & ZoeDepth~\cite{bhat2023zoedepth} &&     9.5  &     1.6  &     3.9  &     8.5  &    14.3  &    19.4  &&    20.4  &     3.0  &     7.5  &    16.7  &    29.7  &    45.3  &&    46.5  &     --  \\ \rowcolor{cyan!16}
MVTracker (ours) & MoGe~\cite{wang2025moge}         &&    13.1  &     2.0  &     4.4  &     9.5  &    19.5  &    30.0  &&    25.0  &     3.4  &     8.0  &    18.4  &    37.5  &    57.6  &&    54.6  &     --  \\ \rowcolor{cyan!16}
MVTracker (ours) & MegaSAM~\cite{li2024_megasam}    && \U{31.6} & \U{5.7}  & \U{14.4} & \U{31.3} & \U{47.4} & \U{59.1} && \U{46.0} & \U{11.0} & \U{25.3} & \U{47.8} & \U{66.8} & \U{79.1} && \U{78.6} &     --  \\
CoTracker3~\cite{karaev24cotracker3} & none required&& \B{64.1} & \B{28.2} & \B{51.3} & \B{72.2} & \B{82.4} & \B{86.4} && \B{77.0} & \B{42.1} & \B{67.1} & \B{85.3} & \B{93.3} & \B{97.0} && \B{91.0} &     --  \\
\midrule
&&&\multicolumn{16}{c}{Our Metrics (where numbers are computed per track and then averaged)} \\
\midrule \rowcolor{cyan!16}
MVTracker (ours) & ZoeDepth~\cite{bhat2023zoedepth} &&    10.3  &     1.8  &     4.4  &     9.5  &    15.6  &    20.4  &&    21.8  &     3.3  &     8.4  &    18.3  &    31.9  &    47.3  &&    46.1  &    59.1 \\ \rowcolor{cyan!16}
MVTracker (ours) & MoGe~\cite{wang2025moge}         &&    14.6  &     2.4  &     5.4  &    11.6  &    22.0  &    31.6  &&    26.7  &     3.9  &     9.2  &    20.5  &    40.4  &    59.7  &&    55.1  &    86.0 \\ \rowcolor{cyan!16}
MVTracker (ours) & MegaSAM~\cite{li2024_megasam}    && \U{35.0} & \U{7.4}  & \U{17.4} & \U{35.7} & \U{51.9} & \U{62.6} && \U{46.4} & \U{12.1} & \U{26.2} & \U{48.1} & \U{66.7} & \U{78.9} && \U{79.8} & \U{14.0} \\
CoTracker3~\cite{karaev24cotracker3} & none required&& \B{66.7} & \B{32.6} & \B{56.2} & \B{74.9} & \B{83.3} & \B{86.4} && \B{77.2} & \B{43.2} & \B{67.6} & \B{85.3} & \B{93.1} & \B{96.7} && \B{91.7} & \B{2.9} \\
\bottomrule
\end{tabular}
\label{tab:tapvid2d}
\end{table*}

\section{Evaluation Metrics} \label{app:subsec:metrics}

There is no established set of metrics for multi-view point tracking in 3D. Thus, we adopt four metrics from prior monocular point-tracking benchmarks~\cite{doersch2023tapvid}, extending them to our multi-view 3D setting. These metrics measure the quality of the predicted 3D trajectories (\davg, MTE), the ability to predict whether a point is visible in any of the views (OA), or both simultaneously (AJ). We compute all metrics per-trajectory before averaging across all tracks in a scene, and then across all scenes in a dataset.

Let $N$ be the number of tracked points, $T$ the number of frames, and $\hat{p}_t^i$ (resp.\ $p_t^i$) the predicted (resp.\ ground-truth) 3D location of track $i$ at time $t$. Similarly, let $\hat{v}_t^i\in\{0,1\}$ and $v_t^i\in\{0,1\}$ be the predicted and ground-truth visibility flags.

\paragraph{Median Trajectory Error (MTE).}
For each track $i$, the median trajectory error measures the median distance between predicted and ground-truth locations over timesteps where the track is visible:
\begin{align}
  MTE_i \;=\; \mathrm{median}\!\Bigl\{\|\hat{p}_t^i - p_t^i\|_2 \;\big|\; t=1,\dots,T,\;v_t^i=1\Bigr\}.
\end{align}
The overall MTE is the mean across all $N$ tracks:
\begin{align}
  MTE \;=\; \tfrac{1}{N}\sum_{i=1}^N MTE_i.
\end{align}

\paragraph{Occlusion Accuracy (OA).}
Occlusion accuracy quantifies how well the model predicts visibility:
\begin{align}
  \mathrm{OA}_i \;=\;& \tfrac{1}{T}\!\sum_{t=1}^T \mathds{1}\!\bigl(\hat{v}_t^i \;=\; v_t^i\bigr),\\
  \mathrm{OA} \;=\;& \tfrac{1}{N}\sum_{i=1}^N \mathrm{OA}_i.
\end{align}

\paragraph{Average Location Accuracy (\texorpdfstring{\davg}{davg}).}
We evaluate the fraction of points within each of $H$ distance thresholds $\{x_1,\dots,x_H\}$ in centimeters. For a single threshold $x$, define the following:
\begin{align}
  \delta_x^i
  \;=\;
  \tfrac{1}{\sum_{t=1}^T v_t^i}
  \sum_{t=1}^T
  \mathds{1}\!\Bigl(v_t^i=1\Bigr)
  \mathds{1}\!\Bigl(\|\hat{p}_t^i - p_t^i\|_2 < x\Bigr)
  .
\end{align}
We then average over all tracks and thresholds:
\begin{align}
  \delta_x
  \;=\;
  \tfrac{1}{N}\sum_{i=1}^N \delta_x^i,
  \quad
  \davg
  \;=\;
  \tfrac{1}{H}\sum_{h=1}^H \delta_{x_h}.
\end{align}

\paragraph{Average Jaccard (AJ).}
This metric jointly assesses spatial and occlusion accuracy at each threshold $x$. Define $\alpha_{t,x}^i = \mathds{1}\!\bigl(\|\hat{p}_t^i - p_t^i\|_2 < x\bigr)$. Then, for track $i$:
\begin{align}
  AJ_x^i
  \;=\;
  \frac{
        \sum_{t=1}^{T}
        v_t^i\hat{v}_t^i\alpha_{t,x}^i
    }{
        \sum_{t=1}^{T}
        \bigl(
            v_t^i
            + (1 - v_t^i)\hat{v}_t^i
            + v_t^i\hat{v}_t^i(1-\alpha_{t,x}^i)
        \bigr)
    }.
\end{align}
Averaging over $N$ tracks yields
\begin{align}
  AJ_x
  \;=\;
  \tfrac{1}{N}\sum_{i=1}^N AJ_x^i,
  \quad
  AJ
  \;=\;
  \tfrac{1}{H}\sum_{h=1}^H AJ_{x_h}.
\end{align}

\subsection{Evaluation Details}

For MV-Kubric and DexYCB, we sample 512 points uniformly from object surfaces and report metrics averaged across static and dynamic points. Panoptic Studio often lacks labeled static objects, so we sample 512 points and compute metrics over all of them. We use four views in the main evaluation tables: for DexYCB, these are the first four calibrated cameras; for Panoptic Studio, we select distant views; and for MV-Kubric, views are randomly sampled per scene. The sensitivity to the selected views is analyzed in \cref{tab:ablate-camera-setups}. Jaccard and location accuracy thresholds are set to 1/2/5/10/20~cm for DexYCB, 0.65/1.3/2.6/5.2/10.4 simulation-scale-adjusted centimeters for MV-Kubric (1 Kubric unit is about 13~cm before simulation-scale adjustment), and 5/10/20/40~cm for Panoptic Studio.

Panoptic Studio trajectories were generated by merging per-view monocular labels from TAPVid-3D~\cite{doersch2023tapvid}, filtered from Dynamic 3DGS~\cite{luiten2023dynamic} 27-view predictions. These labels are often noisy and erroneous, and for example, exhibit point drift (\eg, labels jumping from an elbow to a finger), which is why we use higher evaluation thresholds on this benchmark and recommend interpreting results as well as using the benchmark with caution. MV-Kubric labels are derived from simulated ground-truth trajectories, while DexYCB tracks are extracted from fitted object meshes provided with the dataset. For DexYCB, we sample points on annotated surfaces and track their position using barycentric coordinates of the deforming mesh triangles. Visibility labels on DexYCB can still be noisy due to imperfect Kinect depth maps and segmentation boundaries. The scripts used to process all datasets and generate labels are publicly available in our codebase.

\end{document}